\documentclass{article}
\usepackage{macros}
\usepackage{authblk}

\begin{document}


\title{Natural Reweighted Wake-Sleep}

\author[1]{Csongor~V\'arady} 
\author[2,3]{Riccardo~Volpi}
\author[2,3]{Luigi~Malag\`o}
\author[1]{Nihat~Ay}

\affil[1]{Institute for Data Science Foundations, Hamburg University of Technology, Hamburg, Germany}

\affil[2]{Transylvanian Institute of Neuroscience, Cluj-Napoca, Romania}
\affil[3]{Quaesta AI, Cluj-Napoca, Romania}

\date{}








\maketitle

\begin{abstract}
Helmholtz Machines (HMs) are a class of generative models composed of two Sigmoid Belief Networks (SBNs), acting respectively as an encoder and a decoder. These models are commonly trained using a two-step optimization algorithm called Wake-Sleep (WS) and more recently by improved versions, such as Reweighted Wake-Sleep (RWS) and Bidirectional Helmholtz Machines (BiHM). The locality of the connections in an SBN induces sparsity in the Fisher Information Matrices associated to the probabilistic models, in the form of a finely-grained block-diagonal structure. In this paper we exploit this property to efficiently train SBNs and HMs using the natural gradient. We present a novel algorithm, called Natural Reweighted Wake-Sleep (NRWS), that corresponds to the geometric adaptation of its standard version. In a similar manner, we also introduce Natural Bidirectional Helmholtz Machine (NBiHM). Differently from previous work, we will show how for HMs the natural gradient can be efficiently computed without the need of introducing any approximation in the structure of the Fisher information matrix. The experiments performed on standard datasets from the literature show a consistent improvement of NRWS and NBiHM not only with respect to their non-geometric baselines but also with respect to state-of-the-art training algorithms for HMs. The improvement is quantified both in terms of speed of convergence as well as value of the log-likelihood reached after training. 
\end{abstract}






\section{Introduction}
Deep generative models have been successfully employed in unsupervised learning to model complex and high dimensional distributions thanks to their ability to extract higher-order representations of the data and thus generalize better~\cite{hinton2006fast,MAL-006}. 
An approach which proved to be successful and thus common to several models is based on the use of two separate networks: the  recognition network, i.e., the encoder, which provides a compressed latent representation for the input data, and the generative network, i.e., the decoder, able to reconstruct the observation in output. AutoEncoders (AEs)~\cite{goodfellow2016deep} are a classical example of this paradigm, where both the encoder and the decoder are commonly implemented as deterministic feed-forward networks. Variational AutoEncoders (VAEs)~\cite{kingma2013vae,rezende2014stochastic} introduce an approximate posterior distribution over the latent variables which are then sampled, thus resulting in stochastic networks. In addition, Helmholtz Machines (HMs)~\cite{dayan1995helmholtz} consist of a recognition and a generative network both modelled as Sigmoid Belief Network (SBNs)~\cite{neal1992connectionist}, characterized by discrete hidden variables, differently from standard VAEs which commonly adopt continuous Gaussian variables only in the bottleneck layer.

The training of stochastic networks is a challenging task in deep learning~\cite{glorot2010understanding}. 
This extends to generative models based on stochastic networks, which are commonly trained by the maximization of the likelihood or equivalently by the minimization of a divergence function between the unknown distribution of the data and the one of the generative model. The challenges for the optimization task are due to the presence of terms which are computationally expensive to be estimated, such as the partition function.
A solution to this problem consists in the introduction of a family of tractable approximate posterior distributions, parameterized by the encoder network. In the presence of continuous hidden variables, for which the stochastic back-propagation of the gradient is possible, as in VAEs, the two networks can be trained simultaneously, through the definition of a unique loss function which corresponds to a lower-bound for the likelihood, i.e., the ELBO~\cite{kingma2013vae,rezende2014stochastic}.
In presence of discrete hidden variables, as for HMs, this approach cannot be directly employed, and thus standard training procedures relies on the well-known Wake-Sleep~\cite{hinton1995wake} algorithm, in which two optimization steps for the parameters of the recognition and generative networks are alternated. The Wake-Sleep algorithm, as well as more recent advances~\cite{bornschein_reweighted_2014,bornschein_bidirectional_2015,wenliang2020amortised,hewitt2020learning}, relies on the conditional independence assumption between the hidden variables of each layer, which allows a factorization of the gradient of the loss function associated to directed graphical models~\cite{lauritzen1996}. This leads to a computationally efficient formula for the weights update which does not require the gradients to be back-propagated through the full network. An alternative to Wake-Sleep for HMs is given by the REINFORCE algorithm \cite{williams1992simple}, which is popular in the Reinforcement Learning literature. However, differently from Wake-Sleep, with REINFORCE the variance of the gradient grows linearly with the number of the parameters of the network, an issue addressed in several modern variants~\cite{mnih2014neural,tucker2017rebar,grathwohl2018backpropagation,kool2020estimating}.

Besides the choice of the specific loss function to be optimized, depending on the nature of the generative model, in the literature several approaches to speed-up the convergence during training have been proposed, through the definition of different optimization algorithms. One line of research, initiated by Amari and co-workers~\cite{amari1998natural,amari1997neural}, takes advantage of a geometric framework based on notions of Information Geometry~\cite{amari2000methods}, which leads to the definition of the natural gradient. Whenever the loss function is defined over a statistical manifold of distributions, whose geometry is given by the Fisher-Rao metric, the natural gradient of the function to be optimized corresponds to the Riemannian gradient of the function itself computed with respect to the metric of the manifold.
In general the computation of the natural gradient requires the inversion of the Fisher information matrix (FIM), and for this reason often it cannot be directly applied for the training of large neural network due to its computation cost. Several approaches have been proposed in the literature~\cite{desjardins2013metric,desjardins2015natural,grosse2016kronecker,ollivier2015riemannian,martens2015optimizing}
which are all based on more or less sophisticated approximations of the structure of the FIM. By introducing different forms of independence assumptions between random variables from the network, certain blocks of the FIM are set to zero or alternatively they admit specific representations (such as low-rank updates of a diagonal matrix or Kronecker products of matrices) which allow its efficient inversion. Instead, a different view is provided by Sun and Nielsen~\cite{sun2017relative}, which propose to compute a local version of the Fisher-Rao metric, that they call Relative Fisher Information Metric, used to analyze the local learning dynamics in a large system. Yet a different approach is introduced by Lin et al. \cite{lin2021tractable}, where they describe a  method for the computation of the natural gradient based on the use of local-parameter coordinates, which can be applied to several distributions and algorithms.  
The use of the natural gradient for the training of generative models has been exploited in particular in the works of Lin et. al. Zhang et. al.~\cite{lin2017natural,zhang2017noisy}.
In this paper we follow a different approach for the computation of the natural gradient for the training of a HM which does not require an approximation of the FIM before its empirical evaluation. 

Preliminary results from~\cite{ay2002locality} for the computation of the FIM in directed statistical models, pointed out how the matrix associated with an SBN takes a block-diagonal structure, where the block sizes depend linearly on the size of each hidden layer. This result, which can be seen as a direct consequence of the topology of the directed graphical model associated to the SBN, does not require the introduction of any additional independence assumption between random variables in the FIM. Notice that the level of sparsity for the FIM in SBNs is superior to that associated to the standard assumption of independence between layers~\cite{desjardins2015natural,grosse2016kronecker,ollivier2015riemannian,martens2015optimizing}, where the width of the blocks is given by the product of the sizes of adjacent hidden layers. Indeed for an SBN we have a finely-grained block-diagonal structure for the FIM, with block widths given by the sizes of the hidden layers, which allows a more computationally efficient inversion of the matrix. 

Motivated by these observations we propose efficient geometric adaptations of the Reweighted Wake-Sleep and the Bidirectional Helmholtz Machine, the two best performing algorithms in the literature for the training of HMs, where the gradient is replaced by the corresponding natural gradient.
The intrinsic sparsity of the FIM is a direct consequence of the topology of the two networks composing a HM. As we will show in the paper this has several advantages, above all it allows for an efficient computation of the exact natural gradient for a given mini-batch, without requiring any further assumption on the structure of the FIM.

Our main contributions are the following.
    Firstly, the design of two novel algorithms (NRWS and NBiHM) based on natural gradient for the training of HMs which exploit a finely-grained block-diagonal structure for the FIM. Such structure for the FIM:
a) has never been exploited before in the training of HMs, not even for deterministic networks, as a matter of fact \cite{ay2002locality} does not refer to any application in training;
b) differently from other models it is exact, i.e., for HMs the sparsity structure is not an approximation/assumption but it derives from conditional independence among variables set by the network topology;
c) is made of smaller-sized blocks (thus it is more efficient to be computed) than the standard block-diagonal structure used in previous works~\cite{sun2017relative,zhang2017noisy}.

Secondly, our results on 3 different datasets show that we are able not only to converge to a better value for the loss both in training and test compared to RWS and BiHM (SOA in the literature for HM), but also to achieve faster convergence, both in terms of epochs and wall-clock time. This is a strong result, since natural gradient often suffers from large computation complexity which prevents its use in practice.

The paper is organized as follows. First, in Sections~\ref{sec:HM} and~\ref{sec:WSandRWS} we briefly present the Helmholtz Machine, the Wake-Sleep and the Reweighted Wake-Sleep algorithms. In Sections~\ref{sec:natgrad} and \ref{sec:fisherinfmatrix} we introduce the natural gradient and the FIM, describing its block structure in the case of a HM. In Section~\ref{sec:NRWS} we define the Natural Reweighted Wake-Sleep Algorithm and in Section~\ref{sec:BiHM} we show how with an analogous argument we can compute the natural gradient also for Natural Bidirectional Helmholtz Machine. Finally, in Sections~\ref{sec:experiments} and \ref{sec:conclusions} we discuss our results and draw the conclusions.



\section{Sigmoid Belief Networks and Helmholtz Machines}\label{sec:HM}
Sigmoid Belief Networks (SBNs) \cite{neal1990learning} are a class of models corresponding to a sequence of stochastic layers, which typically consists of vectors of binary random variables. The activations on each layer of an SBN are Sigmoid functions, which generate in output the means of Bernoulli distributions, one for each hidden random variable. 

Let $x$ be the input variables and $h$ the hidden ones, an SBN can be associated to a joint probability distribution $p(x,h)$,
which factorizes as a directed graphical model~\cite{lauritzen1996}
\begin{equation}
     p(x,h) = p(h | x) p(x) =  p(h^{(L)}|h^{({L-1})})  \cdots p(h^{(2)}|h^{(1)}) \, p(h^{(1)}|x) \, p(x) \; , 
\end{equation}
where each random variable in $h^{(i)}$ at layer $i$ only depends on the variables $h^{(i-1)}$ at the previous layer.

\begin{figure}[ht]
\centering
\begin{subfigure}[c]{.47\textwidth}
  \centering
  \includegraphics[width=0.90\textwidth]{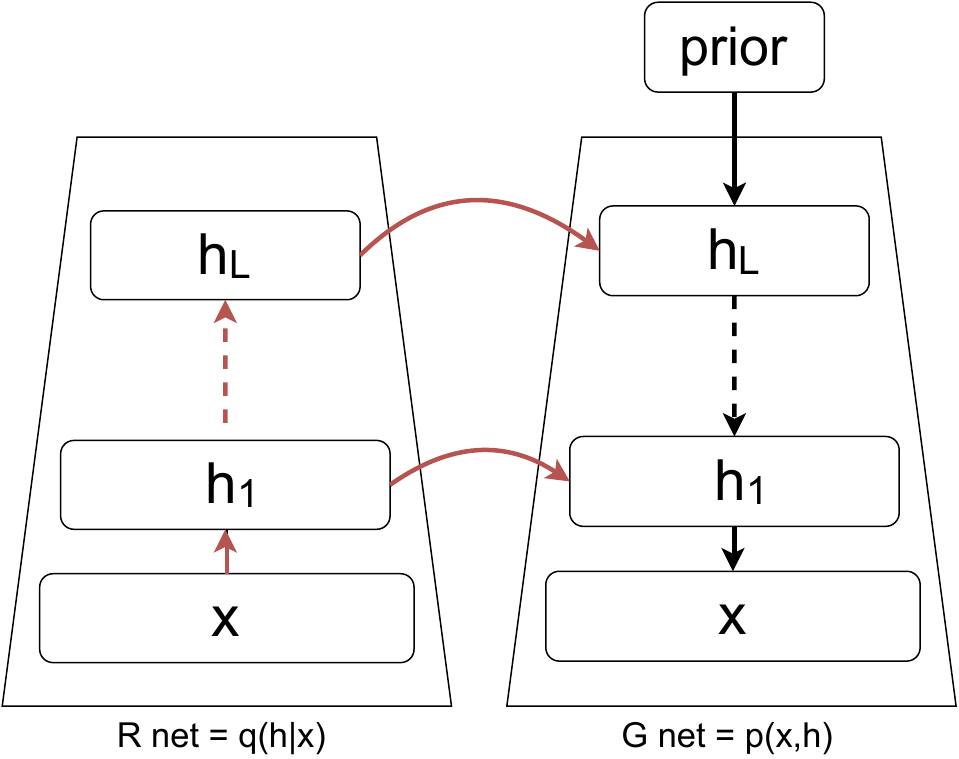}
  \caption{Wake phase}\label{fig:wake}
\end{subfigure}%
\begin{subfigure}[c]{.47\textwidth}
  \centering
  \includegraphics[width=0.90\textwidth]{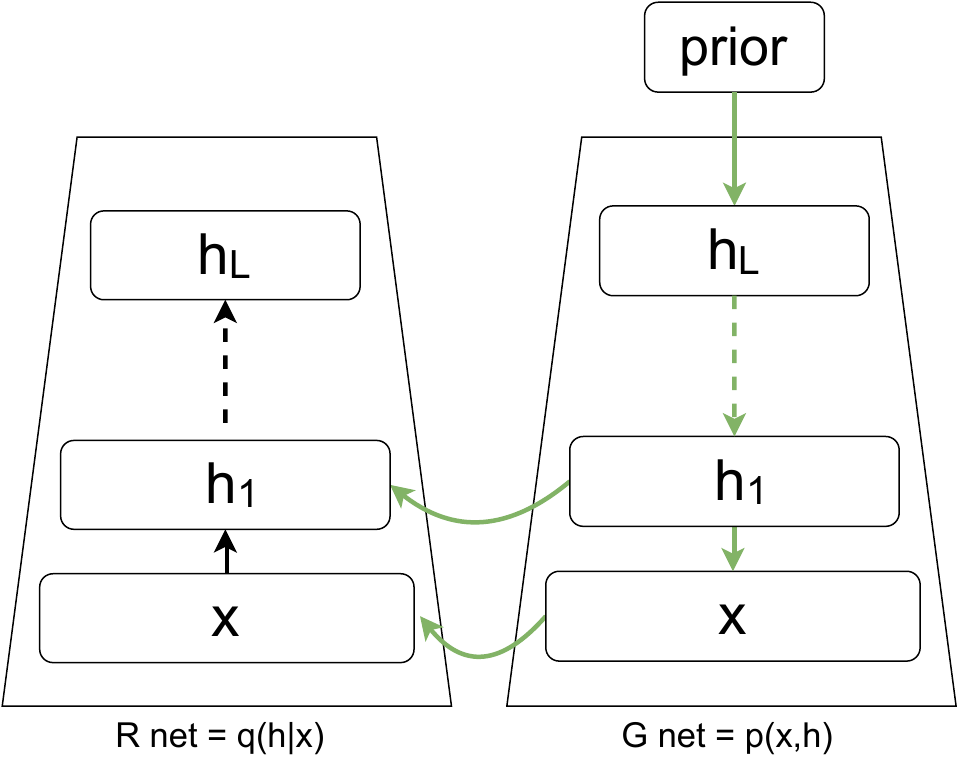}
  \caption{Sleep phase}\label{fig:sleep}
\end{subfigure}
\caption{The structure of a Helmholtz Machine with $L$ layers and a prior distribution over $h^{(L)}$. The colored arrows indicate the propagation of the samples during the Wake \ref{fig:wake} and Sleep \ref{fig:sleep} phases. In the Wake phase the sampling is done from $p_{\Dataset}(x)$ and samples propagate through the recognition network by $q(h|x)$. In the Sleep phase we sample a ``dream'' from the prior $p(h^{(L)})$ and propagate it through the generative network through $p(x|h)$.} \label{fig:HMstruct}
\end{figure}

The Helmholtz Machine (HM) \cite{dayan1995helmholtz} is a generative model which consists of a sequence of layers, one on top of the other, where the layer at the bottom is the visible layer, while the others are the hidden ones. In a HM, the consecutive layers are connected in both directions with two different SBNs. This enables us to define a generative distribution $p$ parameterized by $\theta$ as well as a recognition (conditional) distribution $q$ parameterized by $\phi$. The structure of a HM is illustrated in Figure \ref{fig:HMstruct}. Let $L$ be the number of hidden layers, the distributions $p$ and $q$ factorize as follows:
\begin{align}
    p_\theta(x,h) &= p(x|h^{(1)}) \, p(h^{(1)}|h^{(2)}) \cdots p(h^{(L-1)}|h^{(L)}) \, p(h^{(L)}) \; , \\
    q_\phi(h|x) &= q(h^{(L)}|h^{(L-1)}) \cdots q(h^{(2)}|h^{(1)}) \, q(h^{(1)}|x) \;.
\end{align}
Sometimes we avoid specifying the parametrization $\theta$ and $\phi$ when referring to the distributions of HM for the brevity of equations. However, when the parametrization is missing, it is always assumed that $p$ is parametrized by $\theta$, and $q$ by $\phi$.

Usually in the HM the higher a layer is in the hierarchy the narrower it is in width, with the last layer $L$ being the ``bottleneck'' layer. 

The purpose for a HM is to learn the unknown distribution $p_{\Dataset}(x)$ associated to the observations from a dataset $\Dataset$. Methods to learn such distribution can be obtained by minimizing a divergence function between the parametrized generative distribution $p_\theta(x)$ and $p_{\Dataset}(x)$, for instance
\begin{equation}
 \argmin_\theta \kl{p_{\Dataset}(x)}{p_\theta(x)} \;,
\end{equation}
where 
$\Dif_{KL}$ is the Kullback–Leibler (KL) divergence. If we expand the formula of the KL-divergence, we can easily see that minimizing the divergence is equivalent to minimizing the expected negative log-likelihood of $p(x)$
\begin{equation}\label{eq:loss}
\begin{aligned}
\argmin_\theta &\kl{p_{\Dataset}(x)}{p_\theta (x)} = \argmin_\theta \int p_{\Dataset}(x) \ln \frac{p_{\Dataset}(x)}{p_\theta (x) } dx\\
&= \argmin_\theta  \underbrace{\int p_{\Dataset}(x) \ln p_{\Dataset}(x) dx}_{\text{independent of }\theta} - \int p_{\Dataset}(x) \ln p_\theta (x) dx\\
&= \argmin_\theta  -\mathbb{E}_{p_{\Dataset}(x)} \left[\ln p_\theta (x)\right] \;.
\end{aligned}
\end{equation}
 
Traditionally the HM is trained by the Wake-Sleep algorithm \cite{hinton1995wake} (WS), and the negative log-likelihood $-\ln p_\theta (x)$ is also referred to as the Helmholtz Free Energy. The WS is a two-phase training algorithm, where the wake phase samples from the real distribution $p_{\Dataset}$ and learns the parameters $\theta$ of $p$ by optimizing the Variational Free Energy
\begin{equation}\label{eq:wake}
 E_p(x) = - \ln p_\theta (x) + \kl{q_\phi(h|x)}{p_\theta(h|x)} \;.
\end{equation}
Sometimes in the literature this quantity is also being referred to as evidence lower-bound (ELBO) \cite{kingma2013vae} and it is used as objective function in other optimization methods, cf.~\cite{williams1992simple}.

The sleep phase optimizes the parameters of $q$ based on samples from a prior distribution (called a ``dream'') by taking into account a modified version of the Variational Free Energy, where the arguments of the KL divergence are switched
\begin{equation}\label{eq:sleep}
  E_q(x) = - \ln p_\theta (x) + \underbrace{\kl{p_\theta(h|x)}{q_\phi(h|x)}}_{\text{args switched compared to \eqref{eq:wake}}} \;.
\end{equation}

\section{The Reweighted Wake-Sleep Algorithm}
\label{sec:WSandRWS} 
Following a more recent reinterpretation, the training of Helmholtz Machines can be recast in terms of a variational objective~\cite{bornschein_reweighted_2014, graves2011practical,kingma2013vae}. This is analogous to learning in a Variational AutoEncoder~\cite{kingma2013vae} which requires maximizing a lower bound of the log-likelihood.
Let us start by defining the log-likelihoods for the generation and the recognition probability distributions as 
\begin{align}
    \mathcal{L}_p(x;\theta)&=\ln p_\theta (x)~, \\
    \mathcal{L}_q(x,h;\phi)&=\ln q_\phi (h|x) \;.
\end{align}
For the generation distribution, the derivative of the loss of a single sample $x$ can be estimated as \cite{bornschein_reweighted_2014,le2020revisiting}
\begin{equation}
\begin{aligned}
\label{eq:RLikelihood}
    \frac{\partial \mathcal{L}_p\left(x\sim p_\Dataset(x)\right)}{\partial \theta}
    &= \frac{1}{p(x)} \mathbb{E}_{h\sim q(h|x)}\left[ \frac{p(x,h)}{q(h|x)} \frac{\partial \ln p(x,h)}{\partial \theta} \right]\;.
\end{aligned}
\end{equation}
It is worth noticing that the right hand side of Equation~\eqref{eq:RLikelihood} implies a marginalization over the hidden variables $h$, which marginalization can be approximated via Monte Carlo sampling. During the wake phase a natural choice is made by sampling $h$ from the recognition distribution $q(h|x)$ for the given $x$, i.e. 
\begin{equation}
\label{eq:gradientestimationwake}
\nabla_\theta L_{p}(x) = \sum_{k=1}^S \tilde{\omega}_k \frac{\partial \ln p(x,h^{(k)})}{\partial \theta} \;\; \mbox{with }\;h^{(k)}\sim q(h|x)\;,
\end{equation}
where $S$ is the number of samples considered
and $\nabla_\theta L_{p}(x)$ is defined as the empirical estimate of the wake phase gradient from Equation~\eqref{eq:RLikelihood}. This is called \textbf{p-wake update}. 
The last step is involving the Monte Carlo approximation of the expectation value with importance weights
\begin{equation} \label{eq:weights}
\tilde{\omega}_k = \frac{\omega_k}{\sum_{k'} \omega_{k'}}\;, \mbox{ with  } \; \omega_k = \frac{p(x,h^{(k)})}{q(h^{(k)}|x)} \;.
\end{equation}
The quantity being optimized in Equation~\eqref{eq:RLikelihood} is also referred to as Reconstruction Likelihood (RL) and the optimization is performed in function of the parameters of the generation network $\theta$. 

The approximate posterior $q$ depends on the set of parameters $\phi$, which can be optimized by minimizing the variance of the Monte Carlo estimation in Equation~\eqref{eq:RLikelihood}, or equivalently by minimizing the KL divergence with the generative posterior~\cite{bornschein_reweighted_2014,le2020revisiting}.
This can be averaged by sampling $x$ from the true data distribution $p_{\Dataset}(x)$ (\textbf{q-wake update}) with $h^{(k)} \sim q(h|x)$

\begin{equation} \label{eq:q-wake}
\frac{\partial \mathcal{L}_{q}\left(x\sim p_\Dataset(x)\right)}{\partial \phi} 
\simeq \nabla_\phi L_{q}^{w}(x) = \sum_{k=1}^S \tilde{\omega}_k \frac{\partial \ln q(h^{(k)}|x)}{\partial \phi} \; \mbox{with }\;h^{(k)}\sim q(h|x),
\end{equation}
where $\nabla_\phi L_{q}^{w}(x)$ is empirical estimate of the gradient for the q-wake phase. Alternatively the loss $\mathcal{L}_q$ can be averaged over samples $x,h$ from the generative model (\textbf{q-sleep update}) with $x^{(k)},h^{(k)}\sim p(x,h)$ as
\begin{equation}\label{eq:q-sleep}
\frac{\partial \mathcal{L}_{q}\left((x,h)\sim p(x,h)\right)}{\partial \phi} \simeq  \nabla_\phi L_{q}^{s}(x) = \sum_{k=1}^S \frac{\partial \ln q(h^{(k)}|x^{(k)})}{\partial \phi}\;,
\end{equation}
with $\nabla_\phi L_{q}^{s}(x)$ as the empirical estimate of the gradient for the q-sleep phase.

The Reweighted Wake-Sleep (RWS) \cite{bornschein_reweighted_2014,le2020revisiting} is alternating these three phases during training.
The q-sleep update is commonly known as sleep phase in the classical Wake-Sleep (WS)~\cite{dayan1995helmholtz,hinton1995wake} algorithm, which only uses the p-wake update and the q-sleep update, both with a single sample. 
Indeed with $S=1$, the gradient of the Variational Free Energy in Equation \eqref{eq:wake} with respect to $\theta$ (see e.g., \cite{kirby2006tutorial}) is the same as the gradient of the likelihood in Equation~\eqref{eq:RLikelihood} (see e.g., \cite{bornschein_reweighted_2014} Supplementary Material 6), up to the sign. Moreover, when $S=1$, the gradient with respect to $\phi$ of \eqref{eq:sleep} and \eqref{eq:q-sleep} are obviously the same.

In the following we will simply refer to these phases as wake, q-wake and sleep, which are optimized using the gradients  $\nabla_\theta L_{p}$, $\nabla_\phi L_{q}^{w}$ and $\nabla_\phi L_{q}^{s}$, respectively.

\section{Natural Gradient}
\label{sec:natgrad}
Information Geometry~\cite{amari1985differential,amari2000methods,amari2016information,ay2017information} studies the geometry of statistical models using the language of Riemannian geometry, representing a set of probability distributions $ \mathcal M = \{p_\theta (x) : \theta \in \Theta\}$ as a manifold. Under some regularity conditions, the parametrization $\theta$ for $p$ identifies a set of coordinates, i.e., a chart, over the manifold. Moreover, it is possible to define the tangent space $\tang_p \mathcal M$ in each point $p$ as the set of the velocity vectors along all the curves which pass through $p$.
In Information Geometry, statistical manifolds are commonly endowed with the Riemannian Fisher-Rao metric over the tangent bundle defined by the expected value in $p$ of the product of two tangent vectors, represented by centered random variables. Given a basis for the tangent space, derived from the choice of the parametrization, the inner product associated to the Fisher-Rao metric is represented though a quadratic form given by the Fisher information matrix $\mathcal{F}$. 

Given a real-valued function $\mathcal{L}$ defined over the statistical manifold $\mathcal M$, the direction of steepest ascent is represented by the Riemannian gradient of $\mathcal{L}$ whose evaluation depends on the metric. Let us express $\mathcal{L}(p)$ as function of the parameters $\theta$ by $\mathcal{L}(\theta)$ and let $\nabla \mathcal{L}(\theta)$ denote the vector of partial derivatives $\frac{\partial }{\partial \theta}\mathcal{L}(\theta)$ in the chosen chart.
These are the coordinates of a covector in the cotangent space, i.e., $\frac{\partial }{\partial \theta}\mathcal{L}(\theta) \in \tang_p^{*} \mathcal M$. The natural gradient is the vector in $\tang_p \mathcal M$ associated to $\nabla \mathcal L$ through the canonical isomorphism between tangent and cotangent space induced by the metric~\cite{amari1998natural,amari1997neural}, i.e.,

\begin{equation}\label{eq:NG}
\widetilde{\nabla}\mathcal{L}(\theta) = \mathcal{F}(\theta)^{-1}\nabla\mathcal{L}(\theta)\;,
\end{equation}
with
\begin{equation}
\begin{aligned}
\label{eq:fisher-matrix}
\mathcal{F}(\theta) &= \mathbb E_{p_{\theta}(x)} \left [\frac{\partial}{\partial \theta} \log p_{\theta}(x) \left (\frac{\partial}{\partial \theta} \log p_{\theta}(x) \right)^\trasp \right ] \\
&= - \mathbb E_{p_{\theta}(x)} \left [\frac{\partial^2}{\partial \theta \partial \theta} \log p_{\theta}(x)  \right ]\;.
\end{aligned}
\end{equation}
The natural gradient descent update takes the form of 
\begin{equation}\label{eq:NG2}
    \theta_{t+1} = \theta_{t} - \eta \widetilde{\nabla}\mathcal{L}(\theta_t)\;,
\end{equation}
where $\theta_t$ are the parameters at step $t$ and $\eta > 0$ is the learning rate. 

\section{Fisher Information Matrix for Helmholtz\\ Machines}
\label{sec:fisherinfmatrix}
The computational complexity associated to the evaluation of the Fisher Information Matrix (FIM), needed for the evaluation of the natural gradient of a given loss function, strongly depends on the statistical model on which the loss is defined. We refer the reader to~\cite{park|amari|fukumizu:2000} for a discussion about the evaluation of the FIM for feed-forward networks for classification and regression problems. In this section we show how the FIM in Equation~\eqref{eq:fisher-matrix} can be rewritten in the case of Sigmoid Belief Networks (SBNs), which constitute the building blocks for Helmholtz Machines.
The FIM for directed acyclic graphical models takes a simplified block-diagonal form thanks to the locality of the connection matrix,
given by the conditional independence among the random variables. This result has been exploited recently in the training of stochastic feed-forward neural networks, see for instance Theorem 3 from \cite{sun2017relative}, leading to a block-diagonal FIM with one block per layer, a structure also assumed by \cite{desjardins2015natural,grosse2016kronecker}. 

\begin{figure}[ht!]
\captionsetup[subfigure]{justification=centering}
\centering
\begin{subfigure}[c]{.5\textwidth}
  \centering
  \includegraphics[width=0.95\textwidth]{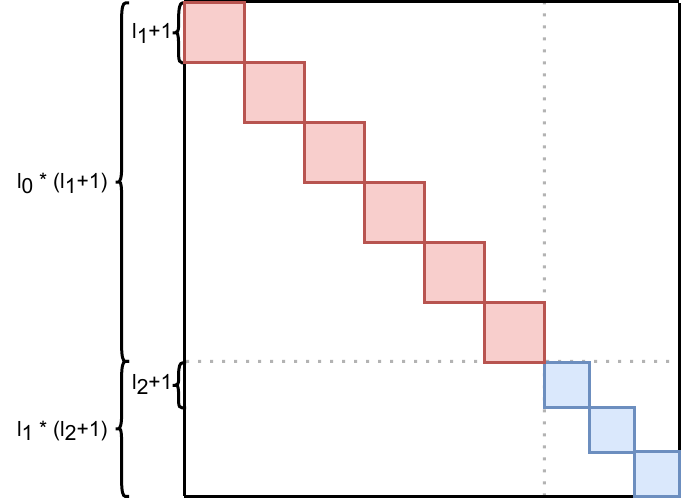}
  \caption{Our structure of \\ the Recognition network FIM}\label{fig:fimRUS}
\end{subfigure}%
\begin{subfigure}[c]{.5\textwidth}
  \centering
  \includegraphics[width=0.8\textwidth]{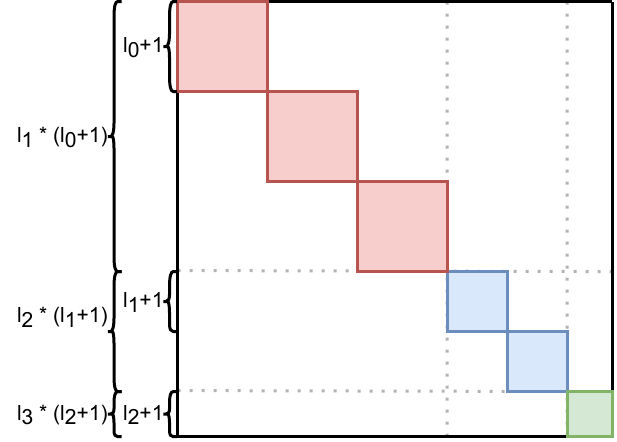}
  \caption{Our structure of \\ the Generation network FIM}\label{fig:fimUS}
\end{subfigure}
\begin{subfigure}[c]{.5\textwidth}
  \centering
  \includegraphics[width=0.95\textwidth]{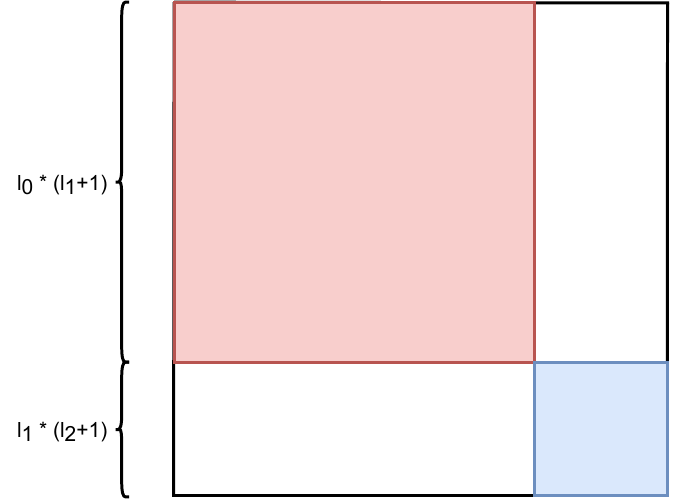}
  \caption{The structure of the Recognition network FIM used in the literature}\label{fig:fimRThem}
\end{subfigure}%
\begin{subfigure}[c]{.5\textwidth}
  \centering
  \includegraphics[width=0.8\textwidth]{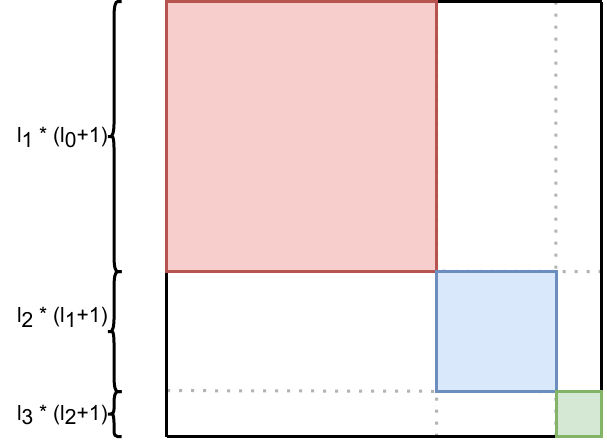}
  \caption{The structure of the Generation network FIM used in the literature}
\end{subfigure}
\caption{(a,b) Graphical representation of the Fisher Information Matrices for the two SBNs in HM with 6-3-2 nodes and a prior distribution on the bottleneck. The gray lines identify the blocks associated to the layers of the network. The matrix admits a fine-grained block-diagonal structure with blocks of size equal to the size of the hidden layers. The blocks are ordered in both cases from the bottom layer to the top. In comparison in (c,d) the block-diagonal FIMs with one block per layer, e.g., ~\cite{desjardins2015natural,grosse2016kronecker,sun2017relative}.}
\label{fig:bStruct}
\end{figure}

However, by generalizing to deep stochastic networks a result from~\cite{ay2002locality} for a two-layers networks, it can be shown that SBNs admit a FIM with a finer-grained block structure, consisting of one block per neuron. This result is a key result for this paper, indeed we can prove that without the need of further approximations, the FIM for SBNs is block-diagonal with blocks of smaller size compared to previous results from the literature, typically having one block per layer, e.g., \cite{desjardins2015natural,grosse2016kronecker,sun2017relative}, which has significant advantages from a computation perspective. The following proposition formalizes this result, while Figure~\ref{fig:bStruct} provides a graphical representation.

\begin{proposition} \label{prop:one} Let $\mathcal G$ be a directed acyclic graphical model, whose variables are grouped in layers such that each node from the $i$-th layer has parent nodes from the $(i-1)$-th layer only. The FIM associated to the joint probability distribution $p$ that factorizes as the product of conditional distributions according to $\mathcal G$ has a block-diagonal structure, with one block for each hidden unit of size equal to the number of parent nodes.
\end{proposition} 
The proof of this result, that we omit here, is based on a generalization of Theorem 1 from \cite{ay2020locality}, see also Lemma 1 in \cite{ay2002locality}, where the locality of natural gradient is studied from a theoretical perspective, without applications to algorithm design.

\subsection{Fisher Information Matrix of a Sigmoid Belief Network}
\label{sec:fishersbn}

Let us consider a Helmholtz Machine with $L$ hidden layers indexed by $i=1, \dots, L$, with binary random vectors $h^{(i)}$, whose distributions are parameterized by $\theta^{(i)}$ which consists of the weights $W^{(i)}$ and biases $b^{(i)}$ at each layer $i$, for $i = 0, .., L$. The $0$-th layer $h^{(0)}=x$ is also called the visible layer. The generation network introduces a prior $p_{
\theta^{(L)}}$ on the top most layer $L$, leading to the factorization 
\begin{equation}
p_{\theta}(x,h) = p_{\theta^{(L)}}(h^{(L)}) \prod_{i=L-1}^0  p_{\theta^{(i)}}(h^{(i)}|h^{(i+1)}) \;,
\end{equation}
%
%
For each neuron $r$ of the layer $i$, $p(h_r^{(i)}|h^{(i+1)})$ is a Bernoulli distribution conditioned on the previous layer
\begin{equation}
\label{eq:bernoullir}
\sigma(W_r^{(i)\trasp} h^{(i+1)} + b_r^{(i)})^{h_r^{(i)}}  \left(1 - \sigma(W_r^{(i)\trasp} h^{(i+1)} + b_r^{(i)})\right)^{1-h_r^{(i)}} 
\end{equation}
where $W_r^{(i)} \in \reals^{l_{i+1}}$ is a column vector of weights (corresponding to a row of $W^{(i)}$) for the neuron $r$ of the layer $i$, and $b_r^{(i)}$ is a bias, while $h^{(i+1)}$ is a column vector.
Equation~\eqref{eq:bernoullir} can be also written in compact form as
\begin{equation}
\label{eq:bernoulliraug}
\sigma(\widetilde{W}_r^{(i)\trasp} \tilde{h}^{(i+1)})^{h_r^{(i)}} \left(1 - \sigma(\widetilde{W}_r^{(i)\trasp} \tilde{h}^{(i+1)})\right)^{1-h_r^{(i)}}\;,
\end{equation}
where $h^{(i+1)}$ and $W_r^{(i)}$ have been augmented with a vector of ones and $b_r^{(i)}$, respectively. However, for brevity and readability purposes we remove the $\sim$ and assume the same structure with the built-in bias.
The second derivative of the log of Equation~\eqref{eq:bernoulliraug} is
\begin{equation}
\frac{\partial^2 }{\partial {W}_{rl}^{(k)}\partial {W}_{rm}^{(k)}} \ln p(h_r^{(i)}|h^{(i+1)}) = -\sigma^{'}\left({W}_r^{(i)\trasp} {h}^{(i+1)}\right){h}_l^{(i+1)} {h}_m^{(i+1)\trasp}\; ,
\end{equation}
where $\sigma$ is the sigmoid function used in SBN, and $\sigma^{'}= \sigma(1 - \sigma)$ is its derivative.

This leads to a natural block-structure for the FIM $\mathcal{F}$ with respect to the weights ${W}_r^{(i)}$ of the distribution $p$ given by
\begin{align}
\mathcal{F}_{p,r}^{(i)}  &= - \mathbb{E}_{p(x,h)}\bigg[\frac{\partial^2 }{\partial {W}_r^{(i)} \partial {W}_r^{(i)\trasp}} \ln p(h_{r}^{(i)}|h^{(i+1)})\bigg]\\
&= \mathbb{E}_{p(x,h)}\left[
\sigma^{'}\left( W^{(i)\trasp}_r {h}^{(i+1)}\right)
{h}^{(i+1)} {h}^{(i+1)\trasp} \right] \;. \label{fisher}
\end{align}

In case we would use $\{\pm 1\}$ for the binary variables instead of $\{0,1\}$, we would have
\begin{equation}
p(h_r^{(i)}|h^{(i+1)}) = \sigma\left(h_r^{(i)} \, \left({W}_r^{(i)\trasp} {h}^{(i+1)}\right)\right)\;,
\end{equation}
while the formula for the the FIM would be the same as in Equation~\eqref{fisher}.

The recognition network targets to approximate the true posterior distribution by 
\begin{equation}
q_\phi(h|x) = \prod_{i=1}^L  q_{\phi^{(i)}}(h^{(i)}|h^{(i-1)}) \;,
\end{equation}
parameterized by $\phi^{(i)}$ with weights $V^{(i)}$ at each layer $i$, for $i = 0, ..., L-1$.
By means of the generative and discriminative networks we can define two different joint distributions over the visible and hidden variables,  $p_{\theta}(x,h)$ and $q_\phi(x,h) = q_\phi(h|x)p_\Dataset(x)$. Both distributions correspond to a statistical manifold for which we are interested in computing the Fisher-Rao metric.

The blocks associated to the $i$-th layer and $j$-th hidden unit, for both $p$ and $q$, read 

\begin{align}
\mathcal{F}^{(i)}_{p,j}  &= \mathbb{E}_{p(x,h)}\bigg[\sigma^{'}
\left( W^{(i)\trasp}_j   h^{(i+1)} \right) \, h^{(i+1)} h^{(i+1)\trasp} \bigg]\; \text{ and} \label{eq:Fisherp} \\
\mathcal{F}^{(i)}_{q,j}  &= \mathbb{E}_{q(x,h)}\bigg[\sigma^{'} \left( V^{(i)\trasp}_j  h^{(i-1)} \right) \, h^{(i-1)} h^{(i-1)\trasp}\bigg]\;. \label{eq:Fisherq}
\end{align} 
$ W^{(i)\trasp}_j$ and $V^{(i)\trasp}_j$ are the transpose of the $j$-th columns of the parameters of the two networks, corresponding to the $j$-th node in the next layer. Therefore the products $ W^{(i)\trasp}_j h^{(i+1)}$ and $ V^{(i)\trasp}_j h^{(i-1)}$ result in scalars. 
The quantities in the expectations are  square matrices obtained from the outer product of two vectors $h h^{\trasp}$ multiplied with a scalar. The sizes of these squared matrices depend on the number of corresponding weights in $W^{(i)}_j$ and $V^{(i)}_j$, respectively, squared, see Figure \ref{fig:bStruct}.

\subsection{Estimation of the Fisher Information Matrix}
\label{sec:fisherEstimation}

Notice that $h^{(i)}$ is a stochastic quantity which can be sampled just based on the values of the nodes from the previous layer, because of the conditional independence. We can use this fact to do a Monte Carlo estimate of each block of $\mathcal F$ with $n$ samples,
Equations~\eqref{eq:Fisherp} and \eqref{eq:Fisherq} can be estimated as 

\begin{align}
F^{(i)}_{p,j}  &= \frac{1}{n} \sum \sigma^{'}\left(W^{(i)\trasp}_j  h^{(i+1)}\right) \, h^{(i+1)} h^{(i+1)\trasp} \\
\label{eq:estimation-Fisherp}
&= H^{(i+1)} Q_{p,j}^{(i)} \left(H^{(i+1)} \right)^\trasp\; \mbox{with }\;h^{(i+1)}\sim p(h^{(i+1)}|h^{(i+2)}) \; \text{ and}\\ 
F^{(i)}_{q,j}  &= \frac{1}{n} \sum \sigma^{'}\left(V^{(i)\trasp}_j h^{(i-1)}\right) \, h^{(i-1)} h^{(i-1)\trasp} \\
\label{eq:estimation-Fisherq}
&= H^{(i-1)} Q_{q,j}^{(i)} \left( H^{(i-1)}\right)^\trasp  \; \mbox{with }\;h^{(i-1)}\sim q(h^{(i-1)}|h^{(i-2)})\;.
\end{align}

In the last step we introduced a matrix representation for the empirical estimation $F$ of $\mathcal F$, where the $H^{(i)}$ matrices are obtained by concatenating for each sample the vector $h^{(i)}$ as a column vector, while the diagonal matrices $Q_{p,j}^{(i)}$ and $Q_{q,j}^{(i)}$ depend on the evaluation of the activation function.
To obtain a lower variance estimation for the expected value in \eqref{eq:Fisherp}, we use samples from the distribution $q(x,h)=q(h|x)p_\Dataset(x)$ and reweigh them using an importance sampling estimator with the same weights $\tilde{\omega}_k$ as in \eqref{eq:weights}, thus we get
\begin{align}
\label{eq:estimation-Fisherp-lowvar}
F^{(i)}_{p,j}  &= \frac{1}{n} \sum \tilde{\omega}_k \sigma^{'}\left(W^{(i)\trasp}_j \tilde{h}^{(i+1)}\right) \, \tilde{h}^{(i+1)} \tilde{h}^{(i+1)\trasp} \\
\label{eq:estimation-Fisherp-lowvar2}
&= \widetilde{H}^{(i+1)} \widetilde{Q}_{p,j}^{(i)} \left(\widetilde{H}^{(i+1)} \right)^\trasp   \; \mbox{with }\;\tilde{h}^{(i+1)}\sim q(\tilde{h}^{(i+1)}|\tilde{h}^{(i)})\;.
\end{align}

Equations~\eqref{eq:estimation-Fisherp}-\eqref{eq:estimation-Fisherp-lowvar} represent the blocks of the empirical Fisher information matrices, for $W^{(i)}_j$ and for $V^{(i)}_j$, respectively. Notice that the empirical estimations in Equations~\eqref{eq:estimation-Fisherp} and \eqref{eq:estimation-Fisherq} are not to be confused with the approximations typically introduced for the simplification of the FIM, needed to make it computationally tractable in feed-forward neural networks. This block structure is very convenient and represents the main argument for the efficiency of the algorithm. 

\begin{rem} \label{rem:anyModel}
The FIMs in Equations~\eqref{eq:Fisherp} and \eqref{eq:Fisherq} only depend on the statistical models associated to the joint distributions $p(x,h)$ and $q(x,h)$, and they are independent from the specific loss function $\mathcal L$, as well as from the chosen training algorithm. Hence, since the model of the Helmholtz Machine remains unchanged, the same FIMs can be used for different training algorithms, such as WS and RWS.
\end{rem}

It is worth mentioning that the FIM of the visible distribution $p(x)$ could also be derived and used for training, which could be better for approximating the real distribution of the data $p_\Dataset(x)$ \cite{ay2020locality}, however it has been suggested~ \cite{ollivier2017information} that the FIM actually profits from the expressivity of the joint distribution $p(x,h)$. Such derivation for the FIM however presents additional complications and its computational feasibility will be explored in future works.

\section{The Natural Reweighted Wake-Sleep Algorithm}
\label{sec:NRWS}
In this section we introduce the Natural Reweighted Wake-Sleep (NRWS) algorithm, a geometric adaptation of the Reweighted Wake-Sleep (RWS) algorithm, where the update of the weights is obtained through the computation of the natural gradient of the different loss functions in the Wake and Sleep updates. 

\subsection{Inversion of the Fisher Information Matrix}\label{ss:inverstion_fisher}

The matrices of the form $H Q H^\trasp$ associated with the estimation of the blocks of the FIM from Equations~\eqref{eq:estimation-Fisherq} and \eqref{eq:estimation-Fisherp-lowvar2} may be singular depending on the number and on the samples in the minibatch used in the estimation compared to the size of the matrices themselves.  
Let $n$ be the size of the minibatch $B$ multiplied by the number of samples $S$ from the network (respectively $p$ or $q$, depending on the FIM under consideration) and $l_i$ the size of the layer $i$. Notice that during training typically $n < l_i$, thus to guarantee the invertibility of the estimated FIM, we add to $H Q H^\trasp$ the identity matrix multiplied by a damping factor $\alpha > 0$ as a form of Tikhonov regularization. Our regularized estimation of the FIM is then

\begin{equation}
\widetilde{F} = \frac{\alpha \identity_l + F}{1+\alpha} \; ,
\end{equation}
so that $\widetilde{F}^{-1} \xrightarrow{} \identity_l$ for $\alpha \xrightarrow{}\infty$ and  $\widetilde{F}^{-1} \xrightarrow{} F^{-1}$ for $\alpha \xrightarrow{} 0$. An experimental analysis for appropriate values for $\alpha$ can be found in Appendix \ref{ss:dp}. 

The computational complexity of the updating rules in Equations~\eqref{eq:finalUpdateRules} is dominated by the matrix inversion. 
In the estimation of the natural gradient, it is more convenient either to invert the matrix itself or to keep in memory its rank-$k$ update representation, depending on the rank of $\widetilde{F}$ and on its size. In presence of a rank-$k$ update representation, we can use the Shermann-Morrison formula to efficiently calculate the inverse of a rank-$k$ update matrix, e.g.~\cite{amari2000adaptive}, by
\begin{equation}\label{eq:fisherInvers}
\begin{aligned}
\widetilde{F}^{-1} &= \left(\frac{\alpha \identity_l +  H Q H^\trasp}{1+\alpha} \right)^{-1} \\&= \frac{1+\alpha}{\alpha}\left( \identity_l -  H(\alpha Q^{-1} +H^\trasp H)^{-1}H^\trasp \right)\;.
\end{aligned}
\end{equation}

For $l_i > n$, by using the Shermann-Morrison formula instead of a straightforward matrix inversion in the larger layers, we can reduce the theoretical computational complexity of the matrix inversion from $\mathcal{O}(l_i^{2.376})$ to $\mathcal{O}(l_i n + n^{2.376})$ for each block. 
In case $l_i<n$, that is for narrower layers at the top of the network, a direct inversion is computed for efficiency. 

The inversion operation for each layer $i$ has to be done $l_{i-1}$ times for each block of $F_p$ and $l_{i+1}$ times for each block of $F_q$. As a consequence, the overall complexity for each update of the NRWS algorithm will be bounded by $\mathcal{O}\left(l_0 \left(l_1 n + n^{2.376}\right)\right)$ where $l_0$ and $l_1$ are the two bottom layers of the Helmholtz Machine, which are usually the largest.

\begin{algorithm}
\DontPrintSemicolon
  Let $x$ be a sample from the dataset \\
  Let $p$ and $q$ be the distributions of the generation and the recognition networks with weights $W$ and $V$ \\
  Let $\tilde{\omega}$ be the importance weights from the RWS  \\
  Let $L$ be the depth of the HM\\
  \CommentSty{\#wake phase update} \\
  \For{each layer $i$ from $q$ ascending with $h^{(0)}$ = $x$}{
    Sample $h^{(i+1)}$ from $q(h^{(i+1)}|h^{(i)})$\\
    Compute the gradients $\nabla_\theta^{(i)}  L_{p}$ with respect to $W^{(i)}$ \\
    Compute the matrices for $(\widetilde{F}_p^{(i)})^{-1}$ for the sub-blocks in $i$ with  $h^{(i+1)}$ and $p(h^{(i)} |h^{(i+1)})$ \\
    $\widetilde{\nabla}_\theta^{(i)}  L_{p}$ = $(\widetilde{F}_p^{(i)})^{-1} \nabla_\theta^{(i)}  L_{p}$ with weights $\tilde{\omega}$\\
    \CommentSty{\#q-wake update} \\
    Calculate $\nabla_\phi^{(i)}  L_{q}^{w}$ and $(\widetilde{F}_q^{(i)})^{-1}$ as in the \texttt{sleep} phase \\ 
    $\widetilde{\nabla}_\phi^{(i)}  L_{q}^{w}$ = $(\widetilde{F}_q^{(i)})^{-1} \nabla_\phi^{(i)}  L_{q}^{w}$ with weights $\tilde{\omega}$\\
    Update $W^{(i)}$ and $V^{(i)}$ with the $\widetilde{\nabla}_\theta^{(i)}  L_{p}$ and $\widetilde{\nabla}_\phi^{(i)}  L_{q}^{w}$\\
    }
    \CommentSty{\#sleep phase update} \\
    \For{each layer $i$ from $p$ descending with $h^{(L)}$ sampled from the prior}{
    Sample $h^{(i-1)}$ from $p(h^{(i-1)}|h^{(i)})$\\
    Compute the gradients $\nabla_\phi^{(i)}  L_{q}^{s}$ with respect to $V^{(i)}$ \\
    Compute the matrices $(\widetilde{F}_q^{(i)})^{-1}$ for the sub-blocks in $i$ with $h^{(i-1)}$ and $q(h^{(i)} |h^{(i-1)})$\\
    $\widetilde{\nabla}_\phi^{(i)}  L_{q}^{s}$ = $(\widetilde{F}_q^{(i)})^{-1} \nabla_\phi^{(i)}  L_{q}^{s}$\\
    Update $V^{(i)}$ with $\widetilde{\nabla}_\phi^{(i)}  L_{q}^{s}$\\
}
\caption{Natural Reweighted Wake-Sleep} \label{alg:classif}
\end{algorithm}

\subsection{K-step update}\label{ss:k-step-update}
Assuming the locality of the gradient descent step update, we can make the assumption that the metric is changing slowly during few training steps. Under this assumption we can reuse the FIM for a certain amount of steps $K$ before recalculating it. We will call this technique the $K$-step update (a similar approach was used in \cite{martens2015optimizing}). When reusing the previously computed FIM blocks, the complexity for each update is dominated by the multiplication of the inverse FIM with the vanilla gradient, and it becomes $\mathcal{O}\left(l_0 \left(l_1 n + n^{2}\right)\right)$, however a consequence is that we are trading memory space for this speed up. Saving the FIM blocks means the memory usage for each layer $l_i$ increases by $\mathcal{O}\left(l_{i+1} n^{2} +l_{i} n\right)$ when using Shermann-Morrison and $\mathcal{O}\left(l_{i+1} l_{i} ^{2}\right)$ using the straightforward inverse for weights $W^{(i)}$ of size $l_{i} \times l_{i+1}$, and analogously for $V^{(i)}$.

Besides the number of samples $S$, the minibatch size $B$ and learning rate $\eta$, two other hyperparameters have been introduced in the NRWS algorithm: the damping factor $\alpha$, needed to invert the estimation of the FIM computed from the samples when it is not full rank, and the number of steps $K$ during which the FIM is frozen, i.e., it is not updated with respect to the new minibatch, for computational efficiency. Hyperparameter tuning for the learning rate $\eta$, the damping factor $\alpha$ and the value for $K$ are presented in the Appendix \ref{sec:appHP}.

Our experiments show that appropriate values for $\alpha$ are in the range $0.01$ to $0.2$, depending on the network topology. The larger the conditioning number of the FIM of the largest layer, or in case the matrix is not full rank, the bigger the difference between $n$ and $max(l_i)$, the larger $\alpha$ should be chosen.  
We also found that $K$ can be kept relatively high with values between $100$ and $1,000$ with almost no loss in performance, but with a significant gain in time. 

This result shows that during training it is possible to avoid to continuously re-estimate the geometry of the manifold of probability distributions, through the estimation of the FIM at each iteration, and that instead a local approximation is sufficient to speed-up the convergence when using the natural gradient. A plausible explanation for this behavior is given by the use of the Tikhonov regularization which allows to obtain more robust estimations for the FIM.

\subsection{Update rules of the Natural Reweighted Wake-Sleep}\label{ss:final_update_rules}

The update rules for the weights $\theta = \left (W^{(1)}, \dots, W^{(L)} \right)$ and $\phi = \left (V^{(1)}, \dots, V^{(L)} \right )$ are given by 


\begin{equation}\label{eq:finalUpdateRules}
\begin{aligned}
    \theta_{t+1} &= \theta_{t} - \eta \widetilde{F}_{p}^{-1} \frac{1}{B} \sum_{r=1}^B \sum_{k=1}^S  \tilde \omega_k  \nabla_\theta L_p^{(k,r)}\;,\\
    \phi_{t+1} &= \phi_{t} - \frac{\eta}{2} \widetilde{F}_{q}^{-1}  \frac{1}{B} \sum_{r=1}^B \sum_{k=1}^S \left(\frac{1}{S}\nabla_\phi L_{q}^{s,(k,r)} + \tilde \omega_k \nabla_\phi L_{q}^{w,(k,r)}\right)\;,
\end{aligned}
\end{equation}
where the gradients of the empirical losses $\nabla_\theta L_p$, $\nabla_\phi L_q^{s}$, and $\nabla_\phi L_{q}^{w}$ 
are computed with minibatches of size $B$, sampled each $S$ times. Notice that, in accordance with the implementation of the RWS algorithm, the learning rate in the updating rule for $\phi$ is halved to average the two gradients. 
In addition, the empirical FIMs $\widetilde{F}_{p}$ and $\widetilde{F}_{q}$ are also estimated with $B$ and $S$, based on Equations~\eqref{eq:estimation-Fisherp}, \eqref{eq:estimation-Fisherq} and \eqref{eq:estimation-Fisherp-lowvar2}, where $n=B\cdot S$.

The overall complexity for each update of the NRWS algorithm is bounded by the sizes of the two bottom layers of the Helmholtz Machine, $l_0$ and $l_1$ , which are usually the largest ones. Every $K$-th step when a new FIM is calculated, the complexity is $\mathcal{O}\left(l_0 \left(l_1 n + n^{2.376}\right)\right)$, while in between steps, when we reuse the FIM, the complexity is $\mathcal{O}\left(l_0 \left(l_1 n + n^{2}\right)\right)$. 

In practice, because we are using a  highly parallelizable programming library for the implementation (see Section~\ref{sec:experiments} and Appendix \ref{sec:fixed_hp}), we can parallelize along the length of the first layer $l_{0}$ with high efficiency, which reduces the complexity further by a factor that depends on the hardware on which the algorithm is run and on the efficiency of the parallelism.

The pseudo-code for NRWS is presented in Algorithm~\ref{alg:classif}. 

\subsection{Convergence Analysis}\label{sec:convergence}

The convergence of the Wake-Sleep algorithm has been studied by Ikeda et al.~\cite{ikeda1999convergence}. In their work the authors show conditions for the theoretical convergence of a modified version of the Wake-Sleep algorithm, identified as a variant of the geometric \textit{em} algorithm. The convergence of the \textit{em} and their relation to the Expectation-Maximization (EM) optimization process is known in literature and in particular has been studied by Fujiwara et al.~\cite{fujiwara1995gradient} and Amari~\cite{amari1995information}.

Ikeda et al.~\cite{ikeda1999convergence} study the convergence of Wake-Sleep first on the factor analysis model.
They point out that the wake-phase is a gradient flow of the m-step. If the WS algorithm ``sleeps well'' by sleeping for multiple steps until convergence, then this is equivalent to the e-step in the em algorithm and thus the procedure converges to the MLE, being equivalent to the Generalized EM algorithm~\cite{mclachlan2007algorithm}.
They subsequently notice how a sufficient condition for the algorithm to work on a general model is that the generative model is realizable by the recognition model, i.e., $p_\theta(x|h) = q_\phi (x|h)$ for some $\theta, \phi$.
Typically however only one step of sleep is performed at each training iteration in the literature, which despite not respecting the convergence guarantees, still it has been found to work efficiently in practice~\cite{dayan1995helmholtz}. 
In Appendix \ref{ss:appSleepWell} we compared the sleep-well algorithm with respect to the standard WS. We showed that taking multiple steps of sleep in one iteration of the algorithm allows a faster convergence, however when the convergence is measured with respect to the elapsed time, standard WS has still an advantage compared the sleep-well algorithm.

Notice that the algorithm by Ikeda et al.~is using the exact FIM, while in the present work we are employing an estimation of the gradients and of the FIM based on the minibatch.
Let us notice that in the training of the model and in the estimation of the FIM RWS and NRWS are using multiple weighted samples for each point in batch, this does not impact on the theoretical convergence properties derived for the WS algorithm, but has the effect to improve the quality of the estimation. Further studies on the convergence properties of RWS and NRWS in relation to the number of samples used in training represents an interesting research direction and will be object of future work. 

\section{Natural Bidirectional Helmholtz Machine}\label{sec:BiHM}

Differently from WS and RWS, the Bidirectional Helmholtz Machine (BiHM) \cite{bornschein_bidirectional_2015}, which obtains better performances compared to the former methods, optimises a lower bound of the log likelihood with respect to the probability distribution 
\begin{equation}\label{eq:bidirectional}
    p^*(x) = \left(\frac{1}{Z} \sqrt{p(x,h)q(x,h)}\right)^2 \;,
\end{equation}
where Z is the normalization constant. The advantage of this method is that both $p$ and $q$ distributions are learned simultaneously without the need for alternating phases. On the other hand, the update rules for BiHM in practice are the same as the \textbf{wake} and \textbf{q-wake} phases from RWS, see Equations~\eqref{eq:RLikelihood} and \eqref{eq:q-wake}, only with different weights $\tilde{\omega}_k$.

Unfortunately, the computation of the FIM for BiHM does not lead to a block-diagonal structure, due to the way in which $p^*$ is defined. However, due to the relationship of the updating rules of BiHM with those of RWS, a possible workaround is to employ in the computation of the natural gradient a block-diagonal matrix with blocks $\widetilde{F}_p$ and $\widetilde{F}_q$. Notice that this is not the FIM for the underlying probability $p^*$ employed by BiHM.
On the other hand, we can consider $p^*$ as a proxy for the computation of the loss of BiHM (see its definition in~\cite{bornschein_bidirectional_2015}) and view the optimization as happening on the manifolds of $p$ and of $q$, rather then on the manifold of $p^*$. Since the inference can always be done in terms of $p$ and $q$, it becomes intuitive to consider the FIMs of $p$ and $q$.

We refer to such algorithm as Natural Bidirectional Helmholtz Machine (NBiHM).



\section{Experiments}
\label{sec:experiments}

For the performance evaluation of NRWS we use the binarized version of the MNIST dataset of handwritten digits \cite{lecun2010mnist} as a standard benchmark. In addition to MNIST, we show the efficacy of the NRWS on the FashionMNIST dataset and a downsampled version of the Toronto Face Dataset (TFD).
In order to test NRWS not only on binary datasets, but continuous ones as well, we resorted to a form of data augmentation, where the gray values of the pixels were taken as probabilities for the visible layer of the HM. In Appendix~\ref{ss:appDataAugment} we provide further details about the experimental evaluation and we present some results on the miniMNIST dataset, a downsampled binarized version of the MNIST dataset, from $28 \times 28$ to $14 \times 14$ (see Figure~\ref{fig:miniMNISTGen} in Appendix \ref{ss:appDataAugment}), similar to the one used by Hinton et. al.~\cite{hinton1995wake}. We use the miniMNIST dataset also for brief explorative analysis of the hyperparameters shown in the Appendix \ref{sec:appHP}, to determine good values for the learning rate, damping factor, and $K$-step parameters.

The reason for the choice of the above mentioned datasets is that they are well studied in the literature and thus provide a perfect first benchmark for our geometric algorithms.
Therefore, we choose the binary MNIST to have benchmarks to compare with, since it is used in both RWS \cite{bornschein_reweighted_2014} and BiHM \cite{bornschein_bidirectional_2015} papers.
Additionally, we consider FashionMNIST and TFD, to evaluate a higher level of complexity for the images while keeping their size constrained (TFD is also used in BiHM \cite{bornschein_bidirectional_2015}).
We do not consider high resolution and color images as they would not be feasible at this stage, since currently we are limited to using densely connected layers and the calculation of the FIM grows quadratically with the layer size (as shown in the complexity analysis in Section~\ref{ss:k-step-update}). 
This is a current practical limitation of NRWS, and will be addressed in future studies (Section~\ref{sec:conclusions}). 

The functions optimized in training differ depending on the algorithm updating phases, see Sections~\ref{sec:WSandRWS} and \ref{sec:BiHM}. To favor comparisons, in our plots we report as loss function the Negative Log-Likelihood (NLL) averaged over minibatches and samples for all algorithms, since the NLL plays a fundamental role in the training of HMs, see Equation \eqref{eq:loss}. 

In addition we compared NRWS also to a version of the algorithm denoted as DNRWS in the experiments, where only the diagonal elements of the FIM are computed and used in the evaluation of the natural gradient. DNRWS employs a rough but common approximation of the FIM which is much faster to invert, and we are interested in assessing whether or not this could be a good trade-off. We give some additional details about DNRWS and its performance in Appendix \ref{s:appDNRWS}. See~\cite{martens2014} for a discussion about how popular training algorithms such as AdaGrad~\cite{duchi2011}, AdaDelta~\cite{zeiler2012}, and Adam~\cite{kingma2014adam} can be interpreted as providing diagonal approximations for the FIM.

Preliminary analysis on the miniMNIST showed a very small standard deviation for the NLL over multiple runs of the same experiment, with different seeds. We tested the miniMNIST dataset with $24$ different seeds and the best hyperparameters (LR $0.002$ and Dp $0.05$ as in Table \ref{tab:soaminiMNIST} and Figure \ref{fig:miniMNISTcurves} in Appendix \ref{ss:appDataAugment}). After $100$ epochs we obtain mean log likelihood of $-28.39$ and std $0.04$ while after $200$ epochs mean $-28.20$ and std $0.03$. This shows that the variance is relatively small for different seeds and gets smaller over time. We repeated the multiple seed experiment on the TFD dataset as well, with the best hyperparameters, with 10 seeds. After $1,000$ epochs, the resulting LL on the test set was mean $-370.0$ and std $0.18$. In the light of these results  we could conclude that the algorithm is robust against randomness and that there is no growing variance problem (usually associated to the REINFORCE algorithm and its variants). Based on these observations we only present a single run per experiment with the confidence that they behave closely to an average run.

\begin{figure}[ht!]
\centering
\begin{subfigure}[b]{1.0\textwidth}
  \centering
  \includegraphics[width=1.0\textwidth]{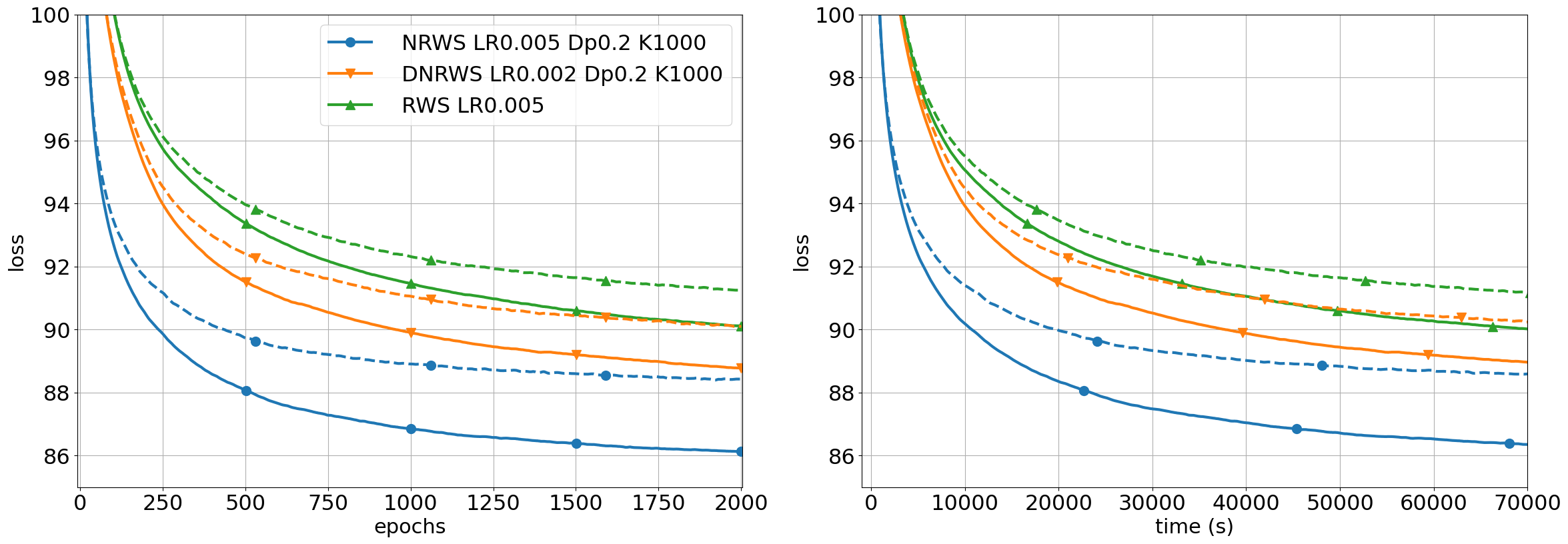}
  \caption{Gradient Descent}\label{fig:soa}
\end{subfigure}
\begin{subfigure}[b]{1.0\textwidth}
  \centering
  \includegraphics[width=1.0\textwidth]{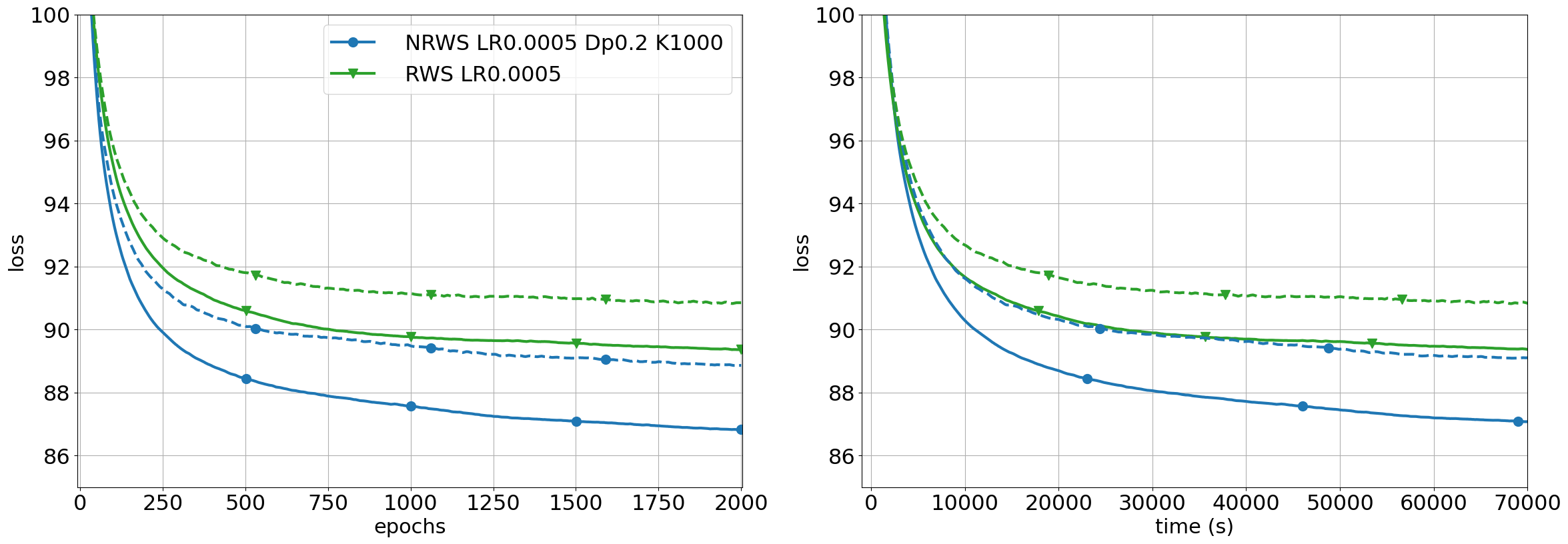}
  \caption{Nadam}\label{fig:soaAdam}
\end{subfigure}%
\caption{Training curves for MNIST for (a) Gradient Descent and (b) Nesterov Adaptive Momentum (Nadam), continuous lines represent the quantities on the train set, and dashed lines the ones on validation. Left: Loss of algorithms over epochs; Right: Loss of algorithms over wall-clock time (s) [LR=learning rate $\eta$, Dp=Damping factor $\alpha$, K=$K$-step].}\label{fig:MNISTGen}
\end{figure}

In all the experiments we worked with an epoch budget and a time budget for the NRWS, or until the algorithm has converged. For Figures~\ref{fig:MNISTGen} and \ref{fig:FashionMNISTGen} we used an epoch budget of $2,000$ and a time budget of $70,000$ seconds which corresponds to roughly 20 hours. 
For each training algorithm, in the plot comparisons we present the results associated to the best choice of the parameters (learning rate $\eta$, $K$, and $\alpha$), optimized for $2,000$ epochs. 

In Appendices \ref{sec:fixed_hp} and \ref{sec:appHP} we report a full description of the experimental settings, source-code, technical information and the complete set of hyperparameters needed to recreate the experiments in this paper.

\subsection{Natural Reweighted Wake-Sleep}\label{ss:exp_nrws} 
We use the model architecture of a binary Helmholtz Machine with layers of sizes 300, 200, 100, 75, 50, 35, 30, 25, 20, 15, 10, 10, as in Bornschein et al.~\cite{bornschein_bidirectional_2015}. The training is performed without data augmentation, with binary variables in $\{-1,1\}$. We used a minibatch size of $B=32$ and a sample size of $S=10$ for all experiments and no regularizers or decaying learning-rate for any of the algorithms.

\subsubsection{MNIST}\label{ss:soa}
In Figure~\ref{fig:soa} and Table~\ref{tab:soa} we report the results of experiments on the MNIST dataset with hyperparameters tuned for each individual algorithm.
The experiments are performed with a binarized dataset, equivalently to other benchmarks in the literature \cite{bornschein_reweighted_2014,bornschein_bidirectional_2015}.
In Figure~\ref{fig:soa} we present the loss curves during training, for the training and validation sets. The advantage of NRWS over RWS in these experiments comes in the form of convergence to a better minimum. NRWS converges faster than its non-geometric counterparts in epochs. 
In time (right panel) NRWS is faster than vanilla RWS, even if the time for each epoch is roughly $25\%$ more. 

\begin{table}[ht!]
\centering
\begin{tabular}{|c|c|c|c|c|c|c|c|}
\hline
 \textbf{ALG} & \textbf{S}  & \boldmath{$\eta$} & \boldmath{$\alpha$} & \textbf{K}& \textbf{LL}  & \textbf{T/E}  \\
   \hline
WS   & 10 & 0.002 & - & - & -90.56 & 30s  \\
RWS  & 10 & 0.002 & - & - & -87.36 & 34s  \\
DNRWS & 10 & 0.002 & 0.2 & - & -86.88 & 39s  \\
NRWS & 10 & 0.002 & 0.2 & 1000 & \textbf{-84.91} & 43s  \\
\hline
VAE \cite{kingma2013vae} & - & - &  - & - & $\approx$ -89.5 & -  \\
RWS \cite{bornschein_reweighted_2014}& 10-100 & {0.001-0.0003} & - & - & $\approx$ -86.0 & -  \\
BiHM \cite{bornschein_bidirectional_2015} & 10-100 & {0.001-0.0003} & - & - & $\approx$ -85.0 & -  \\
\hline
\end{tabular}
\caption{
Importance Sampling estimation of the log-likelihood (\textbf{LL}) on the test set with $10,000$ samples for different algorithms after training till convergence with SGD. \textbf{T/E} is the average time per epoch, \textbf{S} is the number of samples in training, \boldmath$\eta$ is the learning rate, \boldmath$\alpha$ is damping factor and \textbf{K} from $K$-step. The values for VAE, RWS, and BiHM (Bidirectional Helmholtz Machine) are reported from \cite{bornschein_bidirectional_2015} however the \textbf{T/E} are not comparable because of different hardware used in the experiments.}\label{tab:soa}
\end{table}

Natural gradient, by pointing to the steepest direction with respect to the Fisher-Rao metric, allows for higher rates of convergence, however at the same time it might incur in premature convergence and thus reduce generalization properties. This phenomenon is known in the literature and has been already reported by other authors in different contexts~\cite{martens2015optimizing,grosse2016kronecker}. In our experiments we found out that tuning the damping factor was sufficient to regularize the experiments.

We begin by comparing our SGDs implementations of WS, RWS, BiHM, DNRWS, and NRWS with state-of-the-art results from~\cite{bornschein_reweighted_2014,bornschein_bidirectional_2015}. The first section of Table~\ref{tab:soa} presents results associated to our implementations of the algorithms, while the second one reports results from the literature where the training takes advantage of accelerated gradient methods such as ADAM~\cite{kingma2014adam}, learning-rate decay (from $10^{-3}$ to $3\times 10^{-4}$), $L1$ and $L2$ regularizers, and an increased number of samples towards the end of the training (from $10$ to $100$). The use of these techniques lead to improved results compared to plain SGD implementations, however even with a vanilla training procedure adopted in our experiments (fixed learning rate, no regularization and fixed number of samples), we show how the IS Likelihood on $10,000$ samples for NRWS is better than the values reported for RWS from~\cite{bornschein_reweighted_2014} and even slightly better than BiHM~\cite{bornschein_bidirectional_2015}. The impact of variable learning rates and increased number of samples at convergence provides a substantial advantage for BiHM in~\cite{bornschein_bidirectional_2015}, as it can be seen from the results obtained with our implementation discussed in Section~\ref{ss:exp_bid}, where NRWS compares favourably to both BiHM and NBiHM using the same settings in training, up to hyperparameter tuning.
In particular, we expect our results for NRWS to improve further with the use of variable learning rates, additional regularizers, and increased number of samples once the algorithm has reached convergence, as suggested by preliminary results from additional experiments currently in progress.


Additionally, notice that when training until convergence, the difference between the DNRWS and NRWS becomes more significant, we conjecture that the rough approximation of the DNRWS is not able to capture information useful to reach a better optimum. 

Next, we tested the algorithms when training the models using the Nadam (Nesterov-Adam) optimizer \cite{keskar2017improving,wilson2017marginal,kingma2014adam}. In Figure~\ref{fig:soaAdam} we see that also in this case the NRWS benefits from the accelerated gradient method, outperforming RWS both in epochs and in real-world time. We observe that while RWS does seem to be more comparable to NRWS when using Nadam compared to the SGD case, the increase in performance is still not sufficient to catch up to NRWS neither in epochs nor in seconds. The values of the IS Likelihood estimation are reported in Appendix~\ref{s:appRes} and are comparable with those obtained with SGD in Table~\ref{tab:soa}. 

Finally, we observe here that  the adaptive steps and the accumulated momentum of Nadam are computed implicitly assuming an Euclidean geometry for the space of the parameter. However it is known from the literature of accelerated natural gradient that such geometry is not the most convenient one, see e.g.~\cite{chirco|malago|pistone:2021}. This motivates the exploration of adaptive Riemannian gradient methods for the NRWS algorithm, as a future work.

\subsubsection{FashionMNIST and Toronto Face Dataset}

\begin{figure}[ht!]
\centering
\begin{subfigure}[c]{1.0\textwidth}
  \centering
  \includegraphics[width=0.95\textwidth]{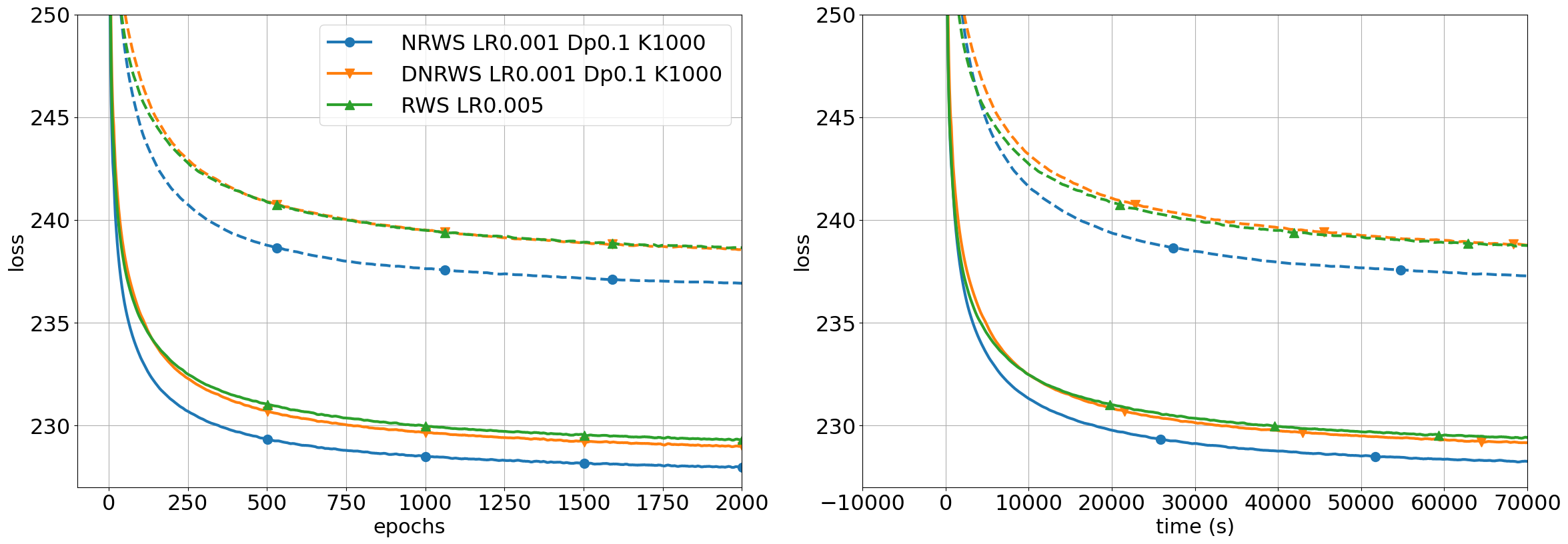}
  \caption{NRWS with gradient descent for FashionMNIST}\label{fig:FIMsoa}
\end{subfigure}
\begin{subfigure}[c]{1.0\textwidth}
  \centering
  \includegraphics[width=0.95\textwidth]{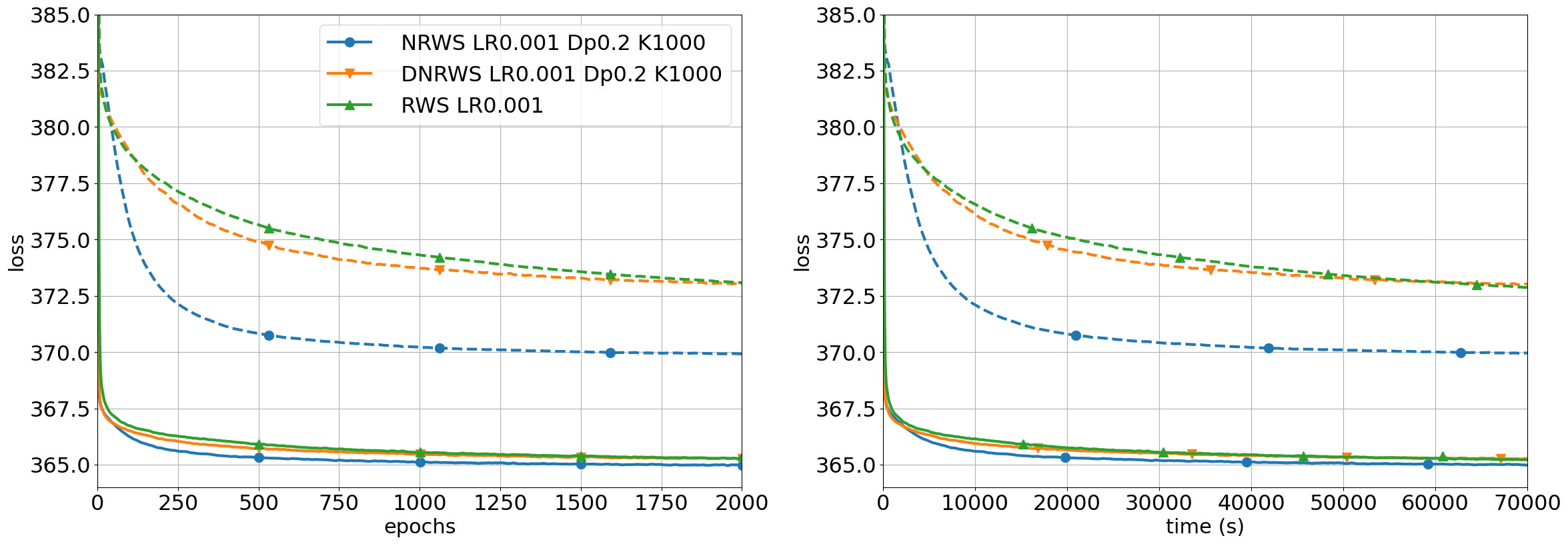}
  \caption{NRWS with gradient descent for TFD}\label{fig:TFDsoa}
\end{subfigure}
\begin{subfigure}[c]{.5\textwidth}
  \centering
  \includegraphics[width=0.95\textwidth]{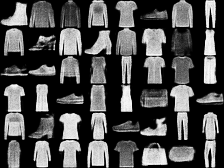}
  \caption{Generated FashionMNIST images}\label{fig:FIMGen}
\end{subfigure}%
\begin{subfigure}[c]{.5\textwidth}
  \centering
  \includegraphics[width=0.95\textwidth]{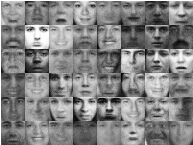}
  \caption{Generated TFD images}\label{fig:TFDGen}
\end{subfigure}%
\caption{Training curves for FashionMNIST (a) and TFD (b) with Gradient Descent, continuous lines represent the quantities on the train set, and dashed lines the ones on validation; Left: Loss of algorithms in epochs; Right: Loss of algorithms in wall-clock time (s) (c) and (d) Example images generated with NRWS after $2,000$ epochs for the FashionMNIST and TFD datasets. [LR=learning rate $\eta$, Dp=Damping factor $\alpha$, K=$K$-step].}\label{fig:FashionMNISTGen}
\end{figure}

We tested NRWS on a downsampled version of the Toronto Face Dataset (TFD) \cite{susskind2010toronto} and on the Fashion MNIST dataset \cite{xiao2017fashion}. We used a $24 \times 24$ resized version for the TFD dataset to be able to use only dense layers.
Given the absence in the literature of experiments on HM with RWS on those datasets, we performed comparisons with our implementation of RWS and NRWS. Our implementation of RWS has been showed to perform as well as the one in the literature, see Table \ref{tab:soa}.

We used a similar setting for experiments for each of the two datasets, as for the MNIST. The same sample and minibatch size and architecture was used for the RWS and NRWS, but the learning rate and damping factor was individually tuned for each algorithm and dataset. For FashionMNIST we used the same network as for the MNIST experiments and for TFD we used 300, 200, 100, 75, 50, 35, 30, 25, 20, 20, which is similar, but wider at the last bottleneck layer.

\begin{table}[ht]
\centering
\begin{tabular}{|c|c|c|c|c|c|c|c|}
\hline
 \textbf{DS} & \textbf{ALG} & \textbf{S}  & \boldmath{$\eta$} & \boldmath{$\alpha$} & \textbf{K} & \textbf{LL}  & \textbf{T/E}  \\
   \hline
\multirow{2}{*}{FashionMNIST} & RWS  & 10 & 0.004 & - & - & -236.96 & 38s  \\
& NRWS & 10 & 0.002 & 0.1 & 1000 & \textbf{-235.65} & 51s  \\
\hline
\multirow{2}{*}{TFD} & RWS  & 10 & 0.002 & - & - & -372.73 & 30s  \\
 & NRWS & 10 & 0.002 & 0.2 & 1000 & \textbf{-370.05} & 39s  \\
\hline
\end{tabular}
\caption{Importance Sampling estimation of the log-likelihood (\textbf{LL}) on the test set with 10,000 samples for different algorithms after training till convergence with SGD. \textbf{T/E} is the average time per epoch, \textbf{S} is the number of samples in training, \boldmath{$\eta$} is the learning rate, \boldmath$\alpha$ is damping factor and \textbf{K} is from $K$-step. The values for RWS are from our own implementation.}\label{tab:soaFMTFD}
\end{table}

In the results from Figure~\ref{fig:FashionMNISTGen} we can observe similar curves to what we saw in the case of the MNIST. Even the best learning rate for the RWS cannot catch up with NRWS neither in epochs nor in real-world time on both datasets. 
In particular, in the case of the TFD dataset we see a large difference in the algorithms ability to generalize. The test curve of the NRWS performs much better than the RWS. In Table \ref{tab:soaFMTFD} we see similar results as for MNIST. NRWS outperforms RWS on both datasets on the test set. In the case of DNRWS the limitations given by the use of a diagonal estimation of the FIM becomes more apparent, as for both datasets the algorithm behaves similarly to RWS. 

\subsection{Natural Bidirectional Helmholtz Machine}\label{ss:exp_bid}
For the experiments with NBiHM and BiHM we use the exact same model architecture, hyperparameters (mini-batch size, sample size) and data-augmentation as we have previously described in Section~\ref{ss:exp_nrws}. Exact values for the hyperparameters that are set on a per-experiment basis (learning-rate, damping factor, $K$-step) are always specified at each experiment where we discuss them and they have been chosen always to favor each algorithm for the given experimental setting.

\subsubsection{MNIST}
\begin{figure}[ht!]
\centering
  \includegraphics[width=1.0\textwidth]{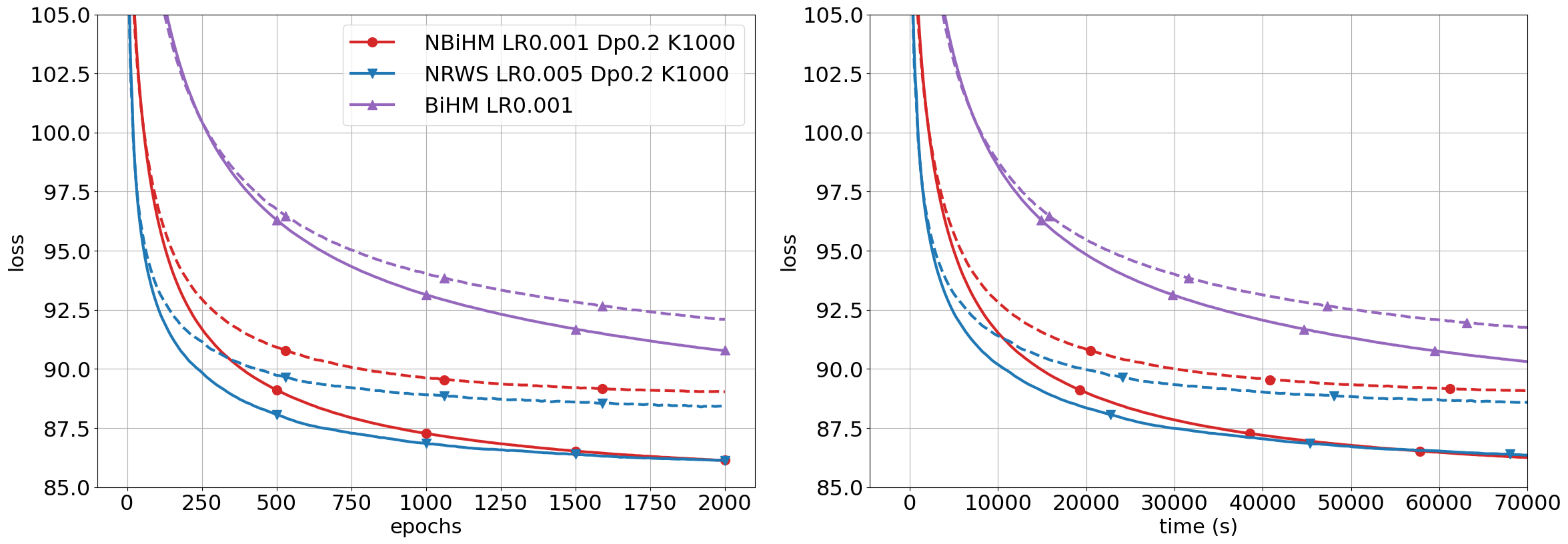}
\caption{Training curves for MNIST for the BiHM and NBiHM algorithms, continuous lines represent the quantities on the train set, and dashed lines the ones on validation. Left: Loss of algorithms over epochs; Right: Loss of algorithms over wall-clock time (s) [LR=learning rate $\eta$, Dp=Damping factor $\alpha$, K=$K$-step].}\label{fig:BiHM}
\end{figure}

We trained BiHM on MNIST as in the original paper~\cite{bornschein_bidirectional_2015}, with Stochastic Gradient Descent with fixed learning rate and sample size, and we compared it to our NBiHM implementation, as in Section~\ref{sec:BiHM}, to evaluate the impact of adapting the computation of the natural gradient for BiHM based on the FIM computed in NRWS. Similarly to our previous experiments we performed them without gradient acceleration, no regularization, and no adaptive sample size. 

In Figure \ref{fig:BiHM}, we notice that NBiHM compared to BiHM benefits from the preconditioning of the gradient with the inverse FIM, both in convergence rate and for the value of the minimum obtained at convergence, even though the FIM is not the proper one, but instead it is the one inherited from RWS. These results are of particular interest, since we are not computing the natural gradient which would correspond to the use of the FIM associated to the $p^*$ from Equation \eqref{eq:bidirectional}.

\begin{table}[ht]
\centering
\begin{tabular}{|c|c|c|c|c|c|c|c|}
\hline
 \textbf{ALG} & \textbf{S}  & \boldmath{$\eta$} & \boldmath{$\alpha$} & \textbf{K} & \textbf{LL $p$} & \textbf{LL $p^*$}  & \textbf{T/E}  \\
   \hline
 BiHM  & 10 & 0.001 & - & - & -87.6 & -90.745 & 29s  \\
 NBiHM & 10 & 0.001 & 0.1 & 1000 & \textbf{-86.18}& \textbf{-89.21} & 38s  \\
\hline
NRWS & 10 & 0.002 & 0.2 & 1000 & \textbf{-84.91} & - & 43s  \\
\hline
\end{tabular}
\caption{Importance Sampling estimation of the log-likelihood (\textbf{LL}) for both $p$ and $p^*$ on the test set for MNIST with 10,000 samples for different algorithms after training till convergence with SGD. \textbf{T/E} is the average time per epoch, \textbf{S} is the number of samples in training, \boldmath{$\eta$} is the learning rate, \boldmath$\alpha$ is damping factor and \textbf{K} is from $K$-step. The values for BiHM and NBiHM are from our own implementation.}\label{tab:soaBiHM}
\end{table}

The same finding can be observed in Table \ref{tab:soaBiHM}. A further observation is that a big advantage of the NBiHM versus the NRWS is its running time. Since the NBiHM is doing two separate updates which are equivalent to the Wake and Wake-q updates up to different reweighting factors, while the NRWS has a total of 3 phases, the NBiHM takes significantly less time for an epoch compared to the NRWS. However, in spite of this, even if NBiHM outperforms BiHM, we could not reach the same accuracy and convergence rate obtained with NRWS, which surpasses both methods. We hypothesize that this could be the side-effect of not using the proper FIM for the algorithm.
  
\subsubsection{FashionMNIST and Toronto Face Dataset}
\begin{figure}[ht!]
\centering
  \includegraphics[width=1.0\textwidth]{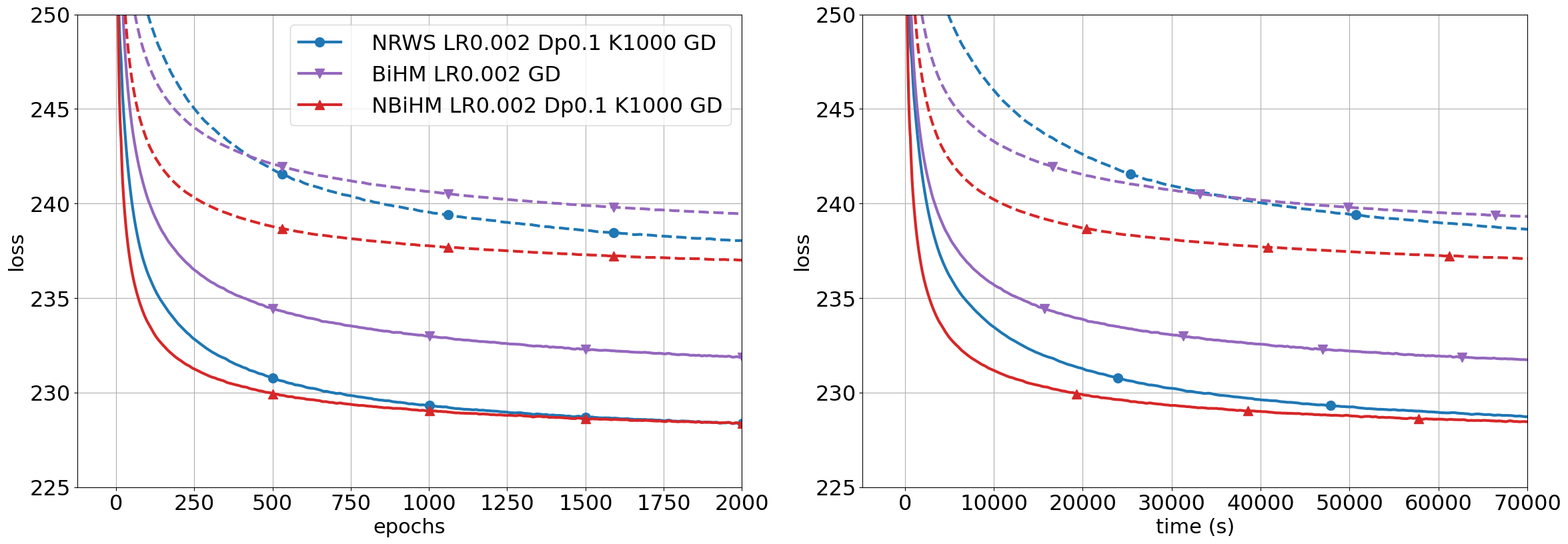}
\caption[NBiHM results on FashionMNIST]{Training curves for FashionMNIST for the BiHM and NBiHM algorithms, continuous lines represent the quantities on the train set, and dashed lines the ones on validation. Left: Loss of algorithms over epochs; Right: Loss of algorithms over wall-clock time (s) [LR=learning rate $\eta$, Dp=Damping factor $\alpha$, K=$K$-step].}\label{fig:BiHM_FashionMNIST}
\end{figure}

\begin{figure}[ht!]
\centering
  \includegraphics[width=1.0\textwidth]{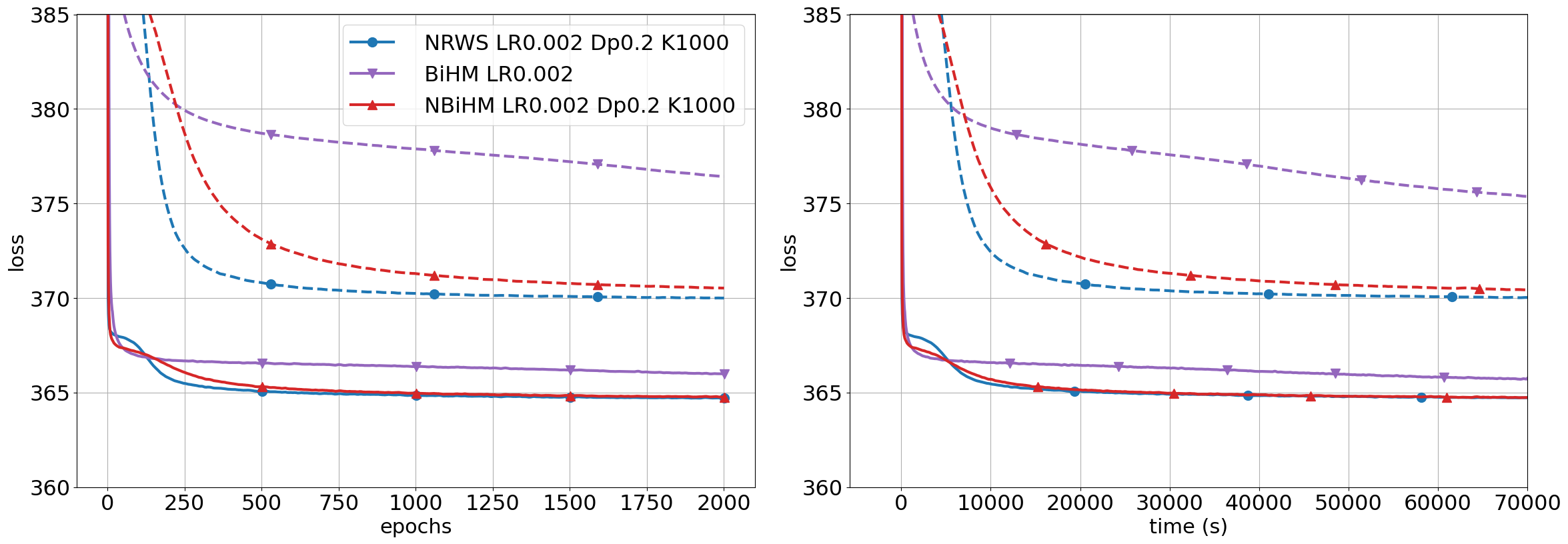}
\caption[NBiHM results on TFD]{Training curves for TFD for the BiHM and NBiHM algorithms, continuous lines represent the quantities on the train set, and dashed lines the ones on validation. Left: Loss of algorithms over epochs; Right: Loss of algorithms over wall-clock time (s) [LR=learning rate $\eta$, Dp=Damping factor $\alpha$, K=$K$-step].}\label{fig:BiHM_TFD}
\end{figure}

\begin{table}[ht]
\centering
\begin{tabular}{|c|c|c|c|c|c|c|c|c|}
\hline
 \textbf{DS} & \textbf{ALG} & \textbf{S}  & \boldmath{$\eta$} & \boldmath{$\alpha$} & \textbf{K} & \textbf{LL} $p$ & \textbf{LL} $p^*$  & \textbf{T/E}  \\
   \hline
\multirow{3}{*}{F-MNIST} & BiHM  & 10 & 0.002 & - & - & -237.99 & -239.41 & 31s  \\
& NBiHM & 10 & 0.002 & 0.1 & 1000 & \textbf{-235.95} & \textbf{-237.15} & 38s  \\
& NRWS & 10 & 0.002 & 0.1 & 1000 & \textbf{-235.65} & - & 48s  \\
\hline 
\multirow{3}{*}{TFD} & BiHM  & 10 & 0.002 & - & - & -375.44 & -375.54 & 27s  \\
 & NBiHM & 10 & 0.002 & 0.2 & 1000 & \textbf{-370.24} & \textbf{-370.39} & 30s  \\
 & NRWS & 10 & 0.002 & 0.2 & 1000 & \textbf{-370.05} & - & 37s  \\
\hline
\end{tabular}
\caption[NBiHM results on FashionMNIST and TFD]{Importance Sampling estimation of the log-likelihood (\textbf{LL}) for both $p$ and $p^*$ on the test set with 10,000 samples for different algorithms after training till convergence with SGD. \textbf{T/E} is the average time per epoch, \textbf{S} is the number of samples in training, \boldmath{$\eta$} is the learning rate, \boldmath$\alpha$ is damping factor and \textbf{K} is from $K$-step.}\label{tab:soaBiHM_TFDFashionMNIST}
\end{table}

We compare BiHM and NBiHM to NRWS on the FashionMNIST and TFD datasets as well, results can be seen in Figures \ref{fig:BiHM_FashionMNIST} and \ref{fig:BiHM_TFD}. The curves on the TFD seem to confirm that NRWS outperforms both NBiHM and BiHM, as previously noticed on the MNIST dataset. On the FashionMNIST dataset instead we observe a different trend, NBiHM is the best overall method, and BiHM keeps a lead on NRWS for half the running time when looking at the wall-clock time. 

The trend previously seen in the training curves is corroborated by the final convergence minima in Table \ref{tab:soaBiHM_TFDFashionMNIST}, where NBiHM shows a large improvement over its non-geometric counterpart, for both log-likelihoods $p$ and $p^*$. However, at convergence, in Table \ref{tab:soaBiHM_TFDFashionMNIST}, the NRWS and NBiHM eventually catch up and achieve for both datasets very close final minima.
Hence, while in the initial phases of training on the FashionMNIST NBiHM prevails, as seen in the training curves, the final values at convergence are approximately within the standard deviation of the experiments (estimated to be approximately $0.18$ on TFD, as shown in the beginning of the present section).


\section{Conclusions}\label{sec:conclusions}
We showed how Helmholtz Machines can be efficiently trained using the natural gradient, thanks to properties of Sigmoid Belief Networks, which allow for an efficient computation of the Fisher information matrix. Indeed by exploiting the locality of the connection matrix given by the network topology of SBNs, the structure of the FIM is a fine grained block-diagonal matrix, finer than what is generally used in the literature. 
In such models it is not required to introduce any extra conditional independence assumption between random variables for computing the natural gradient efficiently, due to the sparse structure of the FIM and use of formulae which allow coarser representations of the blocks in terms of low rank updates of diagonal matrices.

We introduced the Natural Reweighted Wake-Sleep (NRWS) algorithm and we demonstrated an improvement of the convergence during training for stochastic gradient descent. NRWS was not only faster to converge, both in time and number of epochs, but the obtained optimum resulted in better values for the likelihood estimation compared to  RWS~\cite{bornschein_reweighted_2014} and BiHM~\cite{bornschein_bidirectional_2015}.
Our findings have been corroborated by experiments on MNIST as well as on continuous datasets such as FashionMNIST and Toronto Face Dataset. On these datasets, NRWS outperformed the vanilla RWS version in both convergence speed and values of the reached optimum, while also exhibiting a better generalization gap. These results define new state of the art performance for HM on these datasets, not only with respect to WS and RWS, but also in comparison to the more recent BiHM.

Based on the results from Section~\ref{sec:experiments}, we can state that depending on the problem NBiHM represents a viable alternative to NRWS.
Notice that for the NBiHM case, the optimization is defined in terms of $p^*$ (see Section \ref{sec:BiHM}) but the FIM is computed with respect to its components $p$ and $q$. We argue that the computation of the FIM for $p^*$ could benefit further the NBiHM algorithm, but since such a computation is non-trivial, this will be the subject of future work.

The $K$-step update version of the NRWS algorithm showed considerable speed-up in terms of training time, without a decrease in performance, with respect to its baseline with $K=1$. Noticeably, we showed how in our experiments a delayed and thus less accurate estimation of the FIM was sufficient to achieve state-of-the-art performances for HMs. The damping factor introduced in training effectively acts as a regularizer reducing the gap between train and validation.
Studying the effects of regularization over the weights based on the FIM, time-dependent damping factor, $K$-step and learning rate, in order to further boost performances will be object of future studies.

The biggest limitation of the study is currently the scalability of the NRWS and NBiHM, since we focused on dense networks. This limits the sizes and types of datasets on which the algorithm can be efficiently employed. As detailed in Section~\ref{sec:NRWS} the computational complexity of the algorithm is dominated by $\mathcal{O}\left(l_0 \left(l_1 n + n^{2.376}\right)\right)$ where $l_0$ and $l_1$ are the bottom two layers of the Helmholtz Machine, which are usually the largest ones, and $n=B\cdot S$ is minibatch size times sample size. Notice that the computation of the natural gradient requires to define a strategy to store the blocks of the FIM, and thus it has increased memory usage compared to the vanilla gradient.
The current version of the algorithm is designed for dense SBNs and is not suitable for datasets with higher resolution and color images due to the computational complexity depending on $l_0 \cdot l_1$ in the formula above. Nevertheless for such types of datasets more efficient networks based on convolutional filters, using a reduced number of parameters, are usually employed. Hence the scaling limitations might also be seen as a consequence of the use of dense layers.
Further investigations will aim to adapt NRWS to different network topologies, like convolutional networks, while simultaneously employing better estimation techniques further reducing the computational complexity of the algorithm.

We plan to further study ways to obtain more robust estimations of the FIM. Although, as we have shown in the case of the DNRWS, a further simplification of the structure of the FIM can lead to a loss of performance. Additional techniques to refine and accumulate the FIM estimation over time could be beneficial in training.
When using the Nadam optimizer, NRWS seems to maintain the speed advantage and the convergence to a better optimum, compared with its non-geometric counterpart.
This encourages the exploration of adaptive gradient methods for the Natural Reweighted Wake-Sleep in which the Fisher-Rao metric is explicitly considered for the momentum accumulation and the adaptive step.

As a final remark, we highlight that since the computation of the FIM is only dependent on the underlying statistical model, other algorithms for the training of HMs (or in general for the training of network topologies composed of one or more SBNs) could benefit from the use of the natural gradient, as we have already shown for the case of Bidirectional Helmholtz Machine.

\section{Acknowledgements}
V\'arady, Volpi, and Malag\`o have been partially supported by the DeepRiemann project, co-funded by the European Regional Development Fund and the Romanian Government through the Competitiveness Operational Program 2014-2020, Action 1.1.4, project ID P\_37\_714, contract no. 136/27.09.2016.
\bibliographystyle{elsarticle-num} 
\bibliography{references,data_references}

\begin{thebibliography}{10}
\expandafter\ifx\csname url\endcsname\relax
  \def\url#1{\texttt{#1}}\fi
\expandafter\ifx\csname urlprefix\endcsname\relax\def\urlprefix{URL }\fi
\expandafter\ifx\csname href\endcsname\relax
  \def\href#1#2{#2} \def\path#1{#1}\fi

\bibitem{hinton2006fast}
G.~E. Hinton, S.~Osindero, Y.-W. Teh, A fast learning algorithm for deep belief
  nets, Neural computation 18~(7) (2006) 1527--1554.

\bibitem{MAL-006}
Y.~Bengio, Learning {Deep} {Architectures} for {AI}, Foundations and Trends®
  in Machine Learning 2~(1) (2009) 1--127.

\bibitem{goodfellow2016deep}
I.~Goodfellow, Y.~Bengio, A.~Courville, Deep learning, MIT press, 2016.

\bibitem{kingma2013vae}
D.~P. Kingma, M.~Welling, Auto-encoding variational bayes, International
  Conference on Learning Representations - ICLR (2014).

\bibitem{rezende2014stochastic}
D.~J. Rezende, S.~Mohamed, D.~Wierstra, Stochastic backpropagation and
  approximate inference in deep generative models, in: E.~P. Xing, T.~Jebara
  (Eds.), Proceedings of the 31st International Conference on Machine Learning,
  Vol.~32 of Proceedings of Machine Learning Research, PMLR, Bejing, China,
  2014, pp. 1278--1286.

\bibitem{dayan1995helmholtz}
P.~Dayan, G.~E. Hinton, R.~M. Neal, R.~S. Zemel, The {H}elmholtz {M}achine,
  Neural computation 7~(5) (1995) 889--904.

\bibitem{neal1992connectionist}
R.~M. Neal, Connectionist learning of belief networks, Artificial intelligence
  56~(1) (1992) 71--113.

\bibitem{glorot2010understanding}
X.~Glorot, Y.~Bengio, Understanding the difficulty of training deep feedforward
  neural networks, in: Proceedings of the thirteenth international conference
  on artificial intelligence and statistics, 2010, pp. 249--256.

\bibitem{hinton1995wake}
G.~E. Hinton, P.~Dayan, B.~J. Frey, R.~M. Neal, The "wake-sleep" algorithm for
  unsupervised neural networks, Science 268~(5214) (1995) 1158--1161.

\bibitem{bornschein_reweighted_2014}
J.~Bornschein, Y.~Bengio, Reweighted {Wake}-{Sleep}, International Conference
  on Learning Representations - ICLR (2015).

\bibitem{bornschein_bidirectional_2015}
J.~Bornschein, S.~Shabanian, A.~Fischer, Y.~Bengio, Bidirectional {Helmholtz}
  {Machines}, in: International Conference on Machine Learning, PMLR, 2016, pp.
  2511--2519.

\bibitem{wenliang2020amortised}
L.~Wenliang, T.~Moskovitz, H.~Kanagawa, M.~Sahani, Amortised learning by
  wake-sleep, in: International Conference on Machine Learning, PMLR, 2020, pp.
  10236--10247.

\bibitem{hewitt2020learning}
L.~Hewitt, T.~Anh~Le, J.~Tenenbaum, Learning to learn generative programs with
  {Memoised} {Wake}-{Sleep}, in: J.~Peters, D.~Sontag (Eds.), Proceedings of
  the 36th Conference on Uncertainty in Artificial Intelligence (UAI), Vol. 124
  of Proceedings of Machine Learning Research, PMLR, 2020, pp. 1278--1287.

\bibitem{lauritzen1996}
S.~L. Lauritzen, Graphical Models, Oxford University Press, 1996.

\bibitem{williams1992simple}
R.~J. Williams, Simple statistical gradient-following algorithms for
  connectionist reinforcement learning, Machine learning 8~(3-4) (1992)
  229--256.

\bibitem{mnih2014neural}
A.~Mnih, K.~Gregor, Neural variational inference and learning in belief
  networks, in: International Conference on Machine Learning, PMLR, 2014, pp.
  1791--1799.

\bibitem{tucker2017rebar}
G.~Tucker, A.~Mnih, C.~J. Maddison, D.~Lawson, J.~Sohl-Dickstein, Rebar:
  Low-variance, unbiased gradient estimates for discrete latent variable
  models, in: 31st Conference on Neural Information Processing Systems, 2017.

\bibitem{grathwohl2018backpropagation}
W.~Grathwohl, D.~Choi, Y.~Wu, G.~Roeder, D.~Duvenaud, Backpropagation through
  the void: Optimizing control variates for black-box gradient estimation, in:
  International Conference on Learning Representations, 2018.

\bibitem{kool2020estimating}
W.~Kool, H.~van Hoof, M.~Welling, Estimating gradients for discrete random
  variables by sampling without replacement, in: International Conference on
  Learning Representations, 2020.

\bibitem{amari1998natural}
S.-I. Amari, Natural gradient works efficiently in learning, Neural computation
  10~(2) (1998) 251--276.

\bibitem{amari1997neural}
S.-i. Amari, Neural learning in structured parameter spaces-natural
  {Riemannian} gradient, in: Advances in neural information processing systems,
  1997, pp. 127--133.

\bibitem{amari2000methods}
S.-i. Amari, H.~Nagaoka, Methods of information geometry, Vol. 191, American
  Mathematical Soc., 2000.

\bibitem{desjardins2013metric}
G.~Desjardins, R.~Pascanu, A.~Courville, Y.~Bengio, Metric-free natural
  gradient for joint-training of boltzmann machines, International Conference
  on Learning Representations; (2013).

\bibitem{desjardins2015natural}
G.~Desjardins, K.~Simonyan, R.~Pascanu, et~al., Natural neural networks, in:
  Advances in Neural Information Processing Systems, 2015, pp. 2071--2079.

\bibitem{grosse2016kronecker}
R.~Grosse, J.~Martens, A kronecker-factored approximate fisher matrix for
  convolution layers, in: International Conference on Machine Learning, 2016,
  pp. 573--582.

\bibitem{ollivier2015riemannian}
Y.~Ollivier, Riemannian metrics for neural networks {I}: feedforward networks,
  Information and Inference: A Journal of the IMA 4~(2) (2015) 108--153.

\bibitem{martens2015optimizing}
J.~Martens, R.~Grosse, Optimizing neural networks with kronecker-factored
  approximate curvature, in: International conference on machine learning,
  2015, pp. 2408--2417.

\bibitem{sun2017relative}
K.~Sun, F.~Nielsen, Relative fisher information and natural gradient for
  learning large modular models, in: Proceedings of the 34th International
  Conference on Machine Learning-Volume 70, JMLR. org, 2017, pp. 3289--3298.

\bibitem{lin2021tractable}
W.~Lin, F.~Nielsen, K.~M. Emtiyaz, M.~Schmidt, Tractable structured
  natural-gradient descent using local parameterizations, in: International
  Conference on Machine Learning, PMLR, 2021, pp. 6680--6691.

\bibitem{lin2017natural}
W.~Lin, M.~E. Khan, N.~Hubacher, D.~Nielsen, Natural-gradient stochastic
  variational inference for non-conjugate structured variational autoencoder,
  International Conference on Machine Learning (2017).

\bibitem{zhang2017noisy}
G.~Zhang, S.~Sun, D.~Duvenaud, R.~Grosse, Noisy natural gradient as variational
  inference, in: J.~Dy, A.~Krause (Eds.), Proceedings of the 35th International
  Conference on Machine Learning, Vol.~80 of Proceedings of Machine Learning
  Research, PMLR, 2018, pp. 5852--5861.

\bibitem{ay2002locality}
N.~Ay, Locality of global stochastic interaction in directed acyclic networks,
  Neural Computation 14~(12) (2002) 2959--2980.

\bibitem{neal1990learning}
R.~M. Neal, Learning stochastic feedforward networks, Department of Computer
  Science, University of Toronto 64~(1283) (1990) 1577.

\bibitem{graves2011practical}
A.~Graves, Practical variational inference for neural networks, in: Advances in
  neural information processing systems, 2011, pp. 2348--2356.

\bibitem{le2020revisiting}
T.~A. Le, A.~R. Kosiorek, N.~Siddharth, Y.~W. Teh, F.~Wood, Revisiting
  {Reweighted} {Wake}-{Sleep} for models with stochastic control flow, in:
  Uncertainty in Artificial Intelligence, PMLR, 2020, pp. 1039--1049.

\bibitem{kirby2006tutorial}
K.~G. Kirby, A tutorial on {Helmholtz} {Machines}, Department of Computer
  Science, Northern Kentucky University (2006).

\bibitem{amari1985differential}
S.-i. Amari, Differential-geometrical methods in statistics, Lecture Notes on
  Statistics 28 (1985) 1.

\bibitem{amari2016information}
S.-i. Amari, Information geometry and its applications, Vol. 194, Springer,
  2016.

\bibitem{ay2017information}
N.~Ay, J.~Jost, H.~V{\^a}n~L{\^e}, L.~Schwachh{\"o}fer, Information geometry,
  Vol.~64, Springer, 2017.

\bibitem{park|amari|fukumizu:2000}
H.~Park, S.-I. Amari, K.~Fukumizu, Adaptive natural gradient learning
  algorithms for various stochastic models, Neural Networks 13~(7) (2000) 755
  -- 764.

\bibitem{ay2020locality}
N.~Ay, On the locality of the natural gradient for learning in deep {B}ayesian
  networks, Information Geometry (Nov 2020).

\bibitem{ollivier2017information}
Y.~Ollivier, L.~Arnold, A.~Auger, N.~Hansen, Information-geometric optimization
  algorithms: A unifying picture via invariance principles, Journal of Machine
  Learning Research 18~(18) (2017) 1--65.

\bibitem{amari2000adaptive}
S.-i. Amari, H.~Park, K.~Fukumizu, Adaptive method of realizing natural
  gradient learning for multilayer perceptrons, Neural computation 12~(6)
  (2000) 1399--1409.

\bibitem{ikeda1999convergence}
S.~Ikeda, S.-i. Amari, H.~Nakahara, Convergence of the {Wake}-{Sleep}
  {Algorithm}, in: Advances in neural information processing systems, 1999, pp.
  239--245.

\bibitem{fujiwara1995gradient}
A.~Fujiwara, S.-i. Amari, Gradient systems in view of information geometry,
  Physica D: Nonlinear Phenomena 80~(3) (1995) 317--327.

\bibitem{amari1995information}
S.-I. Amari, Information geometry of the {EM} and em algorithms for neural
  networks, Neural networks 8~(9) (1995) 1379--1408.

\bibitem{mclachlan2007algorithm}
G.~J. McLachlan, T.~Krishnan, The {EM} algorithm and extensions, Vol. 382, John
  Wiley \& Sons, 2007.

\bibitem{lecun2010mnist}
[dataset], Y.~LeCun, C.~Cortes, C.~Burges, {MNIST} handwritten digit database
  (2010).

\bibitem{martens2014}
J.~Martens, New insights and perspectives on the natural gradient method,
  Journal of Machine Learning Research 21~(146) (2020) 1--76.

\bibitem{duchi2011}
J.~Duchi, E.~Hazan, Y.~Singer, Adaptive subgradient methods for online learning
  and stochastic optimization, J. Mach. Learn. Res. 12 (2011) 2121–2159.

\bibitem{zeiler2012}
M.~D. Zeiler, {ADADELTA:} an adaptive learning rate method, CoRR abs/1212.5701
  (2012).

\bibitem{kingma2014adam}
D.~P. Kingma, J.~Ba, Adam: A method for stochastic optimization, International
  Conference on Learning Representations - ICLR (2015).

\bibitem{keskar2017improving}
N.~S. Keskar, R.~Socher, Improving generalization performance by switching from
  {ADAM} to {SGD}, International Conference on Learning Representations - ICLR
  (2017).

\bibitem{wilson2017marginal}
A.~C. Wilson, R.~Roelofs, M.~Stern, N.~Srebro, B.~Recht, The marginal value of
  adaptive gradient methods in machine learning, in: Advances in Neural
  Information Processing Systems, 2017, pp. 4148--4158.

\bibitem{chirco|malago|pistone:2021}
G.~Chirco, L.~Malagò, G.~Pistone, Lagrangian and hamiltonian mechanics for
  probabilities on the statistical manifold, arXiv:2009.09431 (2020).

\bibitem{susskind2010toronto}
[dataset], J.~M. Susskind, A.~K. Anderson, G.~E. Hinton, The {T}oronto {F}ace
  {D}atabase, Department of Computer Science, University of Toronto, Toronto,
  ON, Canada, Tech. Rep 3 (2010).

\bibitem{xiao2017fashion}
[dataset], H.~Xiao, K.~Rasul, R.~Vollgraf, Fashion-mnist: a novel image dataset
  for benchmarking machine learning algorithms, arXiv:1708.07747 (2017).

\bibitem{tensorflow2015-whitepaper}
M.~Abadi, A.~Agarwal, P.~Barham, E.~Brevdo, Z.~Chen, C.~Citro, G.~S. Corrado,
  A.~Davis, J.~Dean, M.~Devin, S.~Ghemawat, I.~Goodfellow, A.~Harp, G.~Irving,
  M.~Isard, Y.~Jia, R.~Jozefowicz, L.~Kaiser, M.~Kudlur, J.~Levenberg,
  D.~Man\'{e}, R.~Monga, S.~Moore, D.~Murray, C.~Olah, M.~Schuster, J.~Shlens,
  B.~Steiner, I.~Sutskever, K.~Talwar, P.~Tucker, V.~Vanhoucke, V.~Vasudevan,
  F.~Vi\'{e}gas, O.~Vinyals, P.~Warden, M.~Wattenberg, M.~Wicke, Y.~Yu,
  X.~Zheng, {TensorFlow}: Large-scale machine learning on heterogeneous
  systems, software available from tensorflow.org (2015).

\bibitem{pmlr-v9-glorot10a}
X.~Glorot, Y.~Bengio, Understanding the difficulty of training deep feedforward
  neural networks, in: Y.~W. Teh, M.~Titterington (Eds.), Proceedings of the
  Thirteenth International Conference on Artificial Intelligence and
  Statistics, Vol.~9 of Proceedings of Machine Learning Research, PMLR, Chia
  Laguna Resort, Sardinia, Italy, 2010, pp. 249--256.

\end{thebibliography}

\clearpage
\newpage


\begin{appendices}

\section{Experimental setting and fixed hyperparameters}\label{sec:fixed_hp}

All experiments were run with CUDA optimized Tensorflow 1.15 \cite{tensorflow2015-whitepaper} on Nvidia GTX1080 Ti GPUs. Example configurations to recreate the experiments performed are available in our implementation, which is publicly available at \url{https://github.com/szokejokepu/natural-rws}.

The following hyperparameters were kept fixed throughout all experiments, since the modification of these affects all experiments in a similar way: minibatch size $B=32$, sample size $S=10$ and no regularizers
or decaying learning-rate were used for any of the algorithms. As initialization for the weights we have used the glorot-normal-initializer \cite{pmlr-v9-glorot10a} and a constant initializer of $-1$ for biases, which are also the settings found in \cite{bornschein_bidirectional_2015} as for the BiHM and NBiHM are somewhat more sensitive to initialization, and for other common initializers they reached sub-optimal results.

Learning rates $\mu$, damping factor $alpha$ and $K$-step are dependent on the given experiment and dataset, so specific values for them can be found in the relative table in the main paper.

For the experiment in Figure \ref{fig:soaAdam} for Nadam we have used the standard parameters $\beta_1=0.9$ and $\beta_2=0.999$, which is the default setting.

\section{Hyperparameter tuning}\label{sec:appHP}

In this section we will present in more details some preliminary experiments and hyperparameters tuning principles. In addition to the datasets used in the paper we used the miniMNIST dataset and the ThreeByThree (denoted in the following as 3by3) synthetic dataset. The 3by3 dataset was introduced by Kirby~\cite{kirby2006tutorial} which consists of a 3by3 grid with vertical and horizontal patterns, represented in Fig.~\ref{fig:kstep1} (left). The advantage of this dataset is that the true $KL$ divergence value between the \textit{generation distribution} $p$ and the \textit{true distribution} of the data $p^*$ can be calculated very precisely, and we don't have to rely solely on the approximation of the log-likelihood by importance sampling which is traditionally used to evaluate generative models. The 3by3 converges very quickly to a minimum, so the rate of convergence can be monitored in steps rather than epochs.
These datasets were mainly used as preliminary results to explore the ranges of the hyperparameters quickly and serve as preliminary comparisons between WS, RWS and NRWS.

\begin{figure}[ht]
  \centering
  \includegraphics[width=0.6\linewidth]{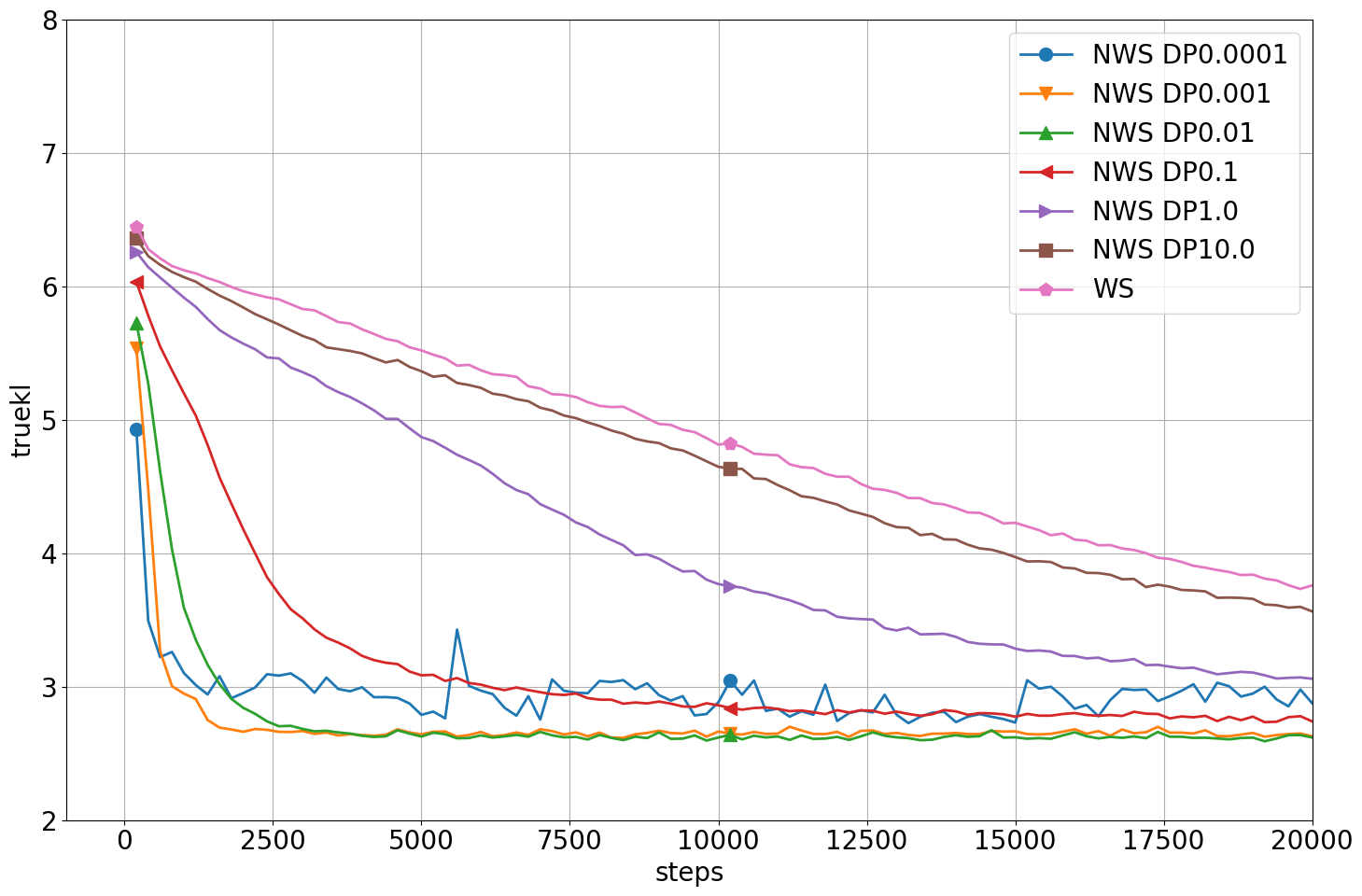}
\caption{The loss with different damping values for the 3by3 dataset. }
  \label{fig:3by3DP}
\end{figure}

\subsection{Learning Rate and Damping Factor}\label{ss:dp}
Ideally we want to find the smallest damping factor which still maintains the optimization stable. Too large damping factors lead to an optimization similar to the non-natural algorithms, while too small damping factors lead to a large conditioning number in the estimated FIM, whose inversion then carries serious numerical issues. This behaviour can be seen clearly in Fig.~\ref{fig:3by3DP}. 
We searched empirically for the right combination of learning rate and damping factor leading to the best convergence rate.

\begin{figure}[ht]
\centering
\includegraphics[width=0.7\textwidth]{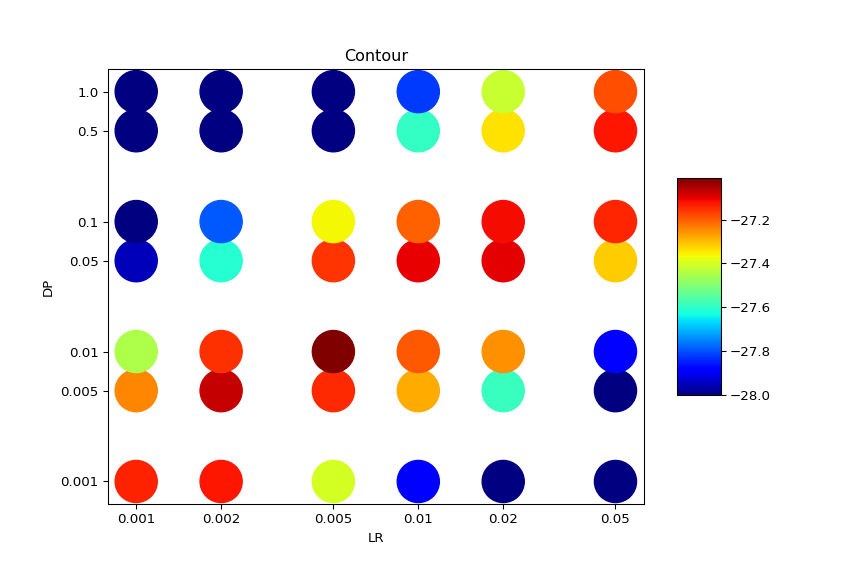}
\caption{The minimum loss of experiments in relation to Learning rate and Damping factor, on a model with 100, 50, 20, 10, 10 layers for miniMNIST.
}
\label{fig:miniMNISTDots}
\end{figure}
\begin{figure}[ht]
\centering
\includegraphics[width=0.7\linewidth]{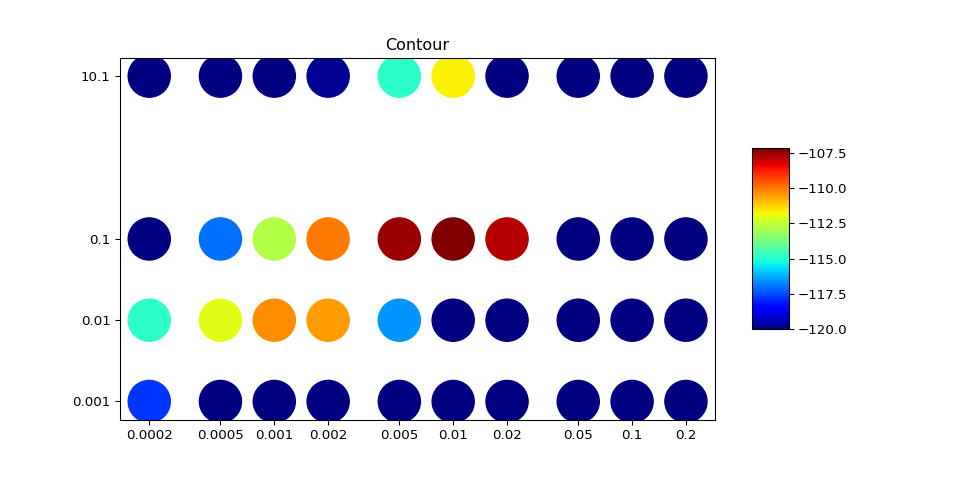}
\caption{The minimum loss of experiments in relation to Learning rate and Damping factor, on a model with 200,100,10 layers for MNIST.}
\label{fig:lrVdp}
\end{figure}

For the 3by3 dataset we determined a range of $0.001 -0.1$ for damping values that outperform the vanilla algorithm (Fig.~\ref{fig:3by3DP}). As one can observe, with very low damping the algorithm converges almost instantly but the loss remains noisy, as each small modification in the FIM is amplified when applied to the gradient. Large values however lead to convergence that is similar to the one of WS. From this general range the parameter has to be fine-tuned for each dataset/model, to determine the appropriate combination of learning rate versus damping factor. 

In Fig.~\ref{fig:miniMNISTDots} and Fig.~\ref{fig:lrVdp} we compare different learning rates and damping factors on a smaller architecture to determine the appropriate quantities for optimal convergence in the case of the miniMNIST and MNIST. We determine that the appropriate damping factor for this dataset with similar models is around $0.005 - 0.1$, with a learning rate in the neighbourhood of $0.005 - 0.02$.

We can notice the expected linear relation between learning rate and damping factor. When we grow the damping factor, we can also grow the learning rate up to a given point. The explanation for this phenomenon is that (as shown in Section~\ref{ss:inverstion_fisher} Inversion of the FIM, in the main paper) the smaller the damping, the closer the algorithm is to following the geometry of the manifold defined by the statistical model. Thus smaller steps lead to better improvements, also a larger dumping mitigates instabilities due to few-samples statistical estimations.

We also observe a correlation between the size of the image and the magnitude of the damping factor. The larger the image in the first layer, the more samples are needed to estimate the FIM of the network accurately. When the number of samples $S$ cannot be grown anymore for practical reasons (the complexity of the algorithm grows quadratically with the number of samples), a larger damping factor is needed so the matrix can have a reasonable conditioning number.

\subsection{K-step}\label{ss:fme}

\begin{figure}[ht!]
\centering 
  \begin{subfigure}[c]{.3\textwidth}
  \centering
  \vspace{-0.3cm}
  \includegraphics[width=0.9\textwidth]{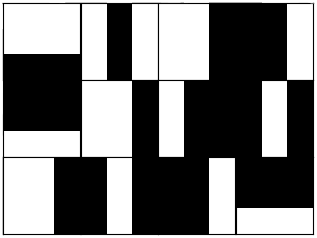}
  \vspace{0.5cm}
  \caption{3by3 samples}
\label{fig:kstep1}
  \end{subfigure}
  \begin{subfigure}[c]{.69\textwidth}
  \centering
\includegraphics[width=\linewidth]{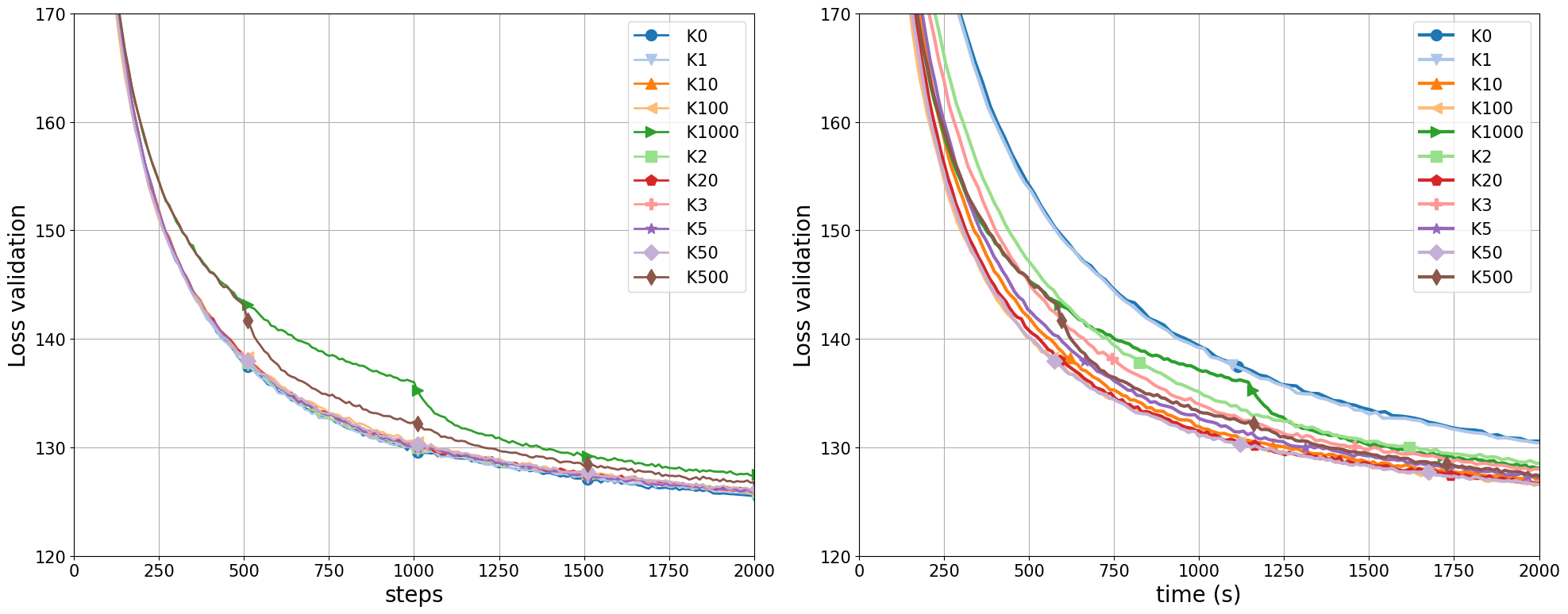}
\caption{Loss on the validation set for MNIST}
\label{fig:kstep2}
\end{subfigure}
\caption{(a) Samples from the 3by3 dataset; (b) Loss curves for MNIST for different values of K=\{0,1,2,3,5,10,20,50,100,500,1000\}; left: over epochs; right: over seconds.}
\end{figure}

Once found the best combination of hyperparameters for the Natural Wake-Sleep optimizer, we explore the possibility to compute the FIM only every K-steps and thus speed up the computational time of the algorithm.
In Fig.~\ref{fig:kstep2} we see the changes in the loss on the validation set for MNIST for a fixed learning rate and damping factor. We observe that for a K-steps the algorithm speeds up significantly, but loses stability at very high values, which is most noticeable in the case of $K=1000$ steps, and to a lesser degree for $K=500$. Consequently we can use a K-step in the range of $[1,100]$ without noticeable impact on the performance of the algorithm. In our experiments we used $K=50$ or $K=100$, keeping in mind that the 3by3 dataset is very simplistic and might not generalize well to other datasets.

\begin{table}
 \begin{minipage}[b]{.5\textwidth}
  \centering
    \includegraphics[width=0.9\textwidth]{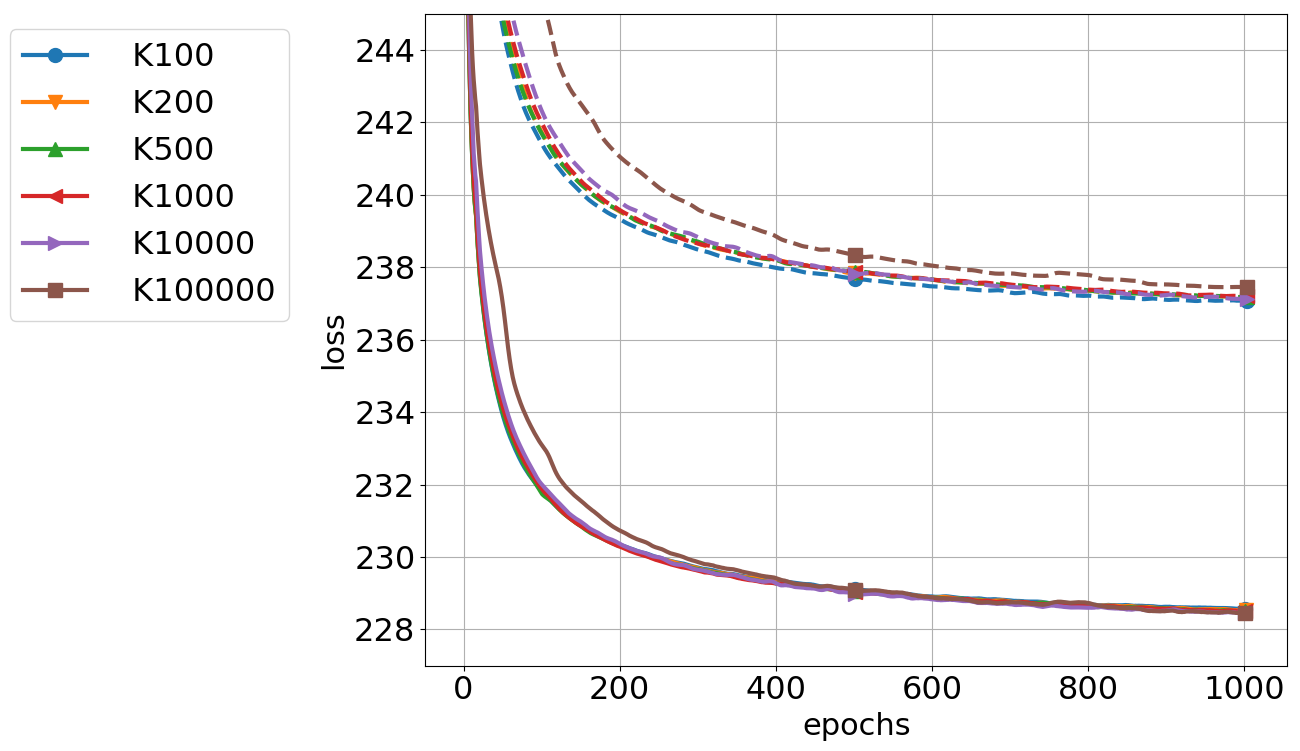}
    \captionof{figure}{Loss curves for train and test on FashionMNIST for different values of $K$.}\label{fig:varyK}
\end{minipage}\hspace{0.5cm}%
\begin{minipage}[b]{.4\textwidth}
    \centering
  \begin{tabular}{|r|r|}
      \hline
      \textbf{K} & \textbf{median. time (s)} \\
      \hline
      100 & 286.17 \\
      200 & 234.74 \\
      500 & 207.63 \\
      1,000 & 201.74 \\
      10,000 & 191.10 \\
      100,000 & 184.28 \\
      \hline
  \end{tabular}
  \vspace{0.5cm}
\caption{Average time per epoch for different values of $K$.}\label{tab:varyK}
\end{minipage}
\end{table}

However, the analysis we did thus far on $K$-step was mainly focused on the short term effects. To thoroughly explore the effects of the K-step, we analyze the convergence of the NRWS on the FashionMNIST dataset, we fix learning rate $0.001$ and damping factor $0.1$ while varying K. In Figure \ref{fig:varyK} we can see that clearly $K=100000$ is pushing the algorithm too far, as it breaks down on both the train and test curves. All-in-all we notice that most of the curves converge to the same minimum, contrary to what we might have expected from the previous experiment. However there is a significant difference in the time an epoch takes in Table \ref{tab:varyK}. 

Furthermore we can notice a slight advantage in the test curve for $K=100$, which hints at keeping the $K$ value as low as possible might still bring some advantage. We conclude that for more complex image datasets, such as FashionMNIST and TFD, $K$ values in the range of $100-1000$ are the most preferable, which are acceptable from a speed and final convergence perspective.
The fact that we can keep the FIM unchanged for so many steps is somewhat surprising and might hint to the fact that for some datasets the metric on the manifold is varying slowly from point to point.


\section{Sleep-well}\label{ss:appSleepWell}
As preliminary exploratory analysis we studied briefly the Sleep-well variant of Wake-Sleep. As mentioned in Section~\ref{sec:convergence} WS only has theoretical convergence guarantees for the variant where the sleep phase is done until convergence after every wake phase. At the best of our knowledge there are no studies in the literature using this variant. It is usually commonly accepted to use WS as a simple alternating algorithm, with one step of each phase instead.

\begin{figure}
\centering
\includegraphics[width=0.8\textwidth]{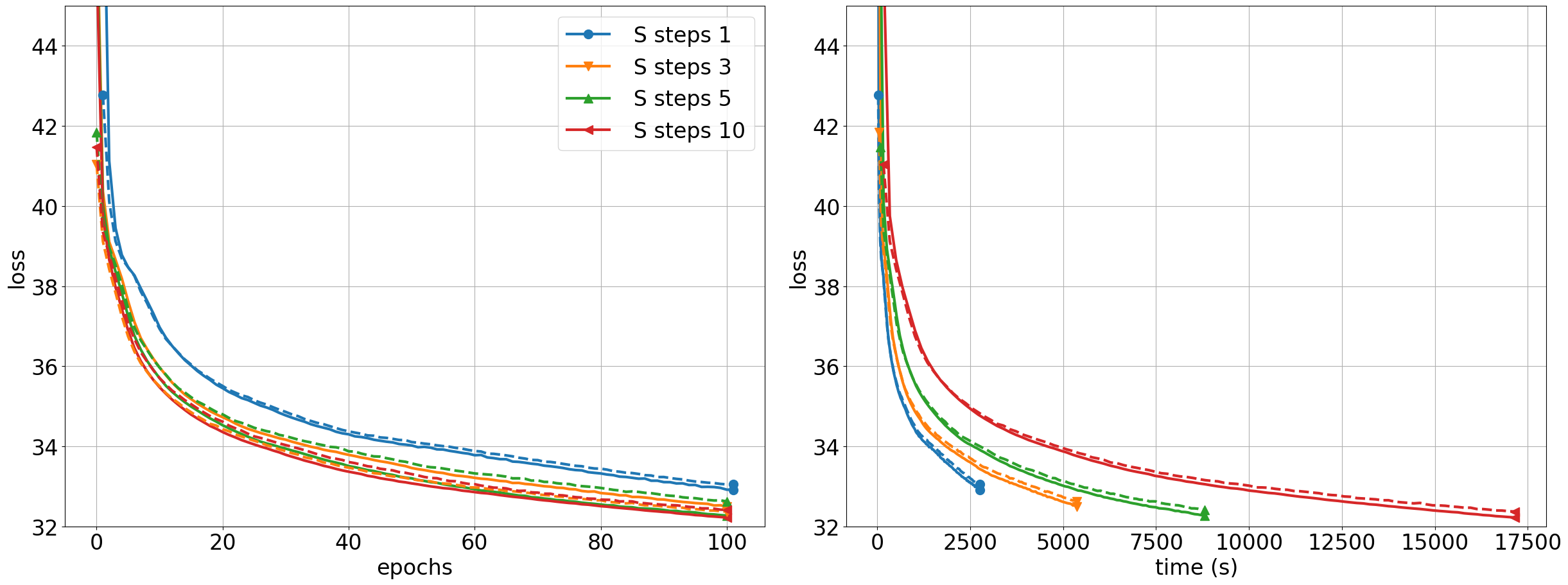}
\caption{Loss on the train and validation sets of the miniMNIST with different 1, 3, 5, 10 sleep steps for every wake step for WS.}\label{fig:SleepWell}
\end{figure}

In Figure~\ref{fig:SleepWell} we compare 4 variants of the WS on miniMNIST, with the same hyperparameters as above, where we only changed the number of sleep steps: 1, 3, 5, 10 per one wake step. We see that there is a noticeable difference between the variants, with more sleep steps resulting in better convergence in epochs. Looking at the real-time comparisons of the experiments, the conclusion takes a different perspective. The time penalty for the multiple sleep phases ends up slowing down the algorithm significantly, with an amount that scales linearly with the number of steps. 


\begin{figure}
\centering
\includegraphics[width=0.8\textwidth]{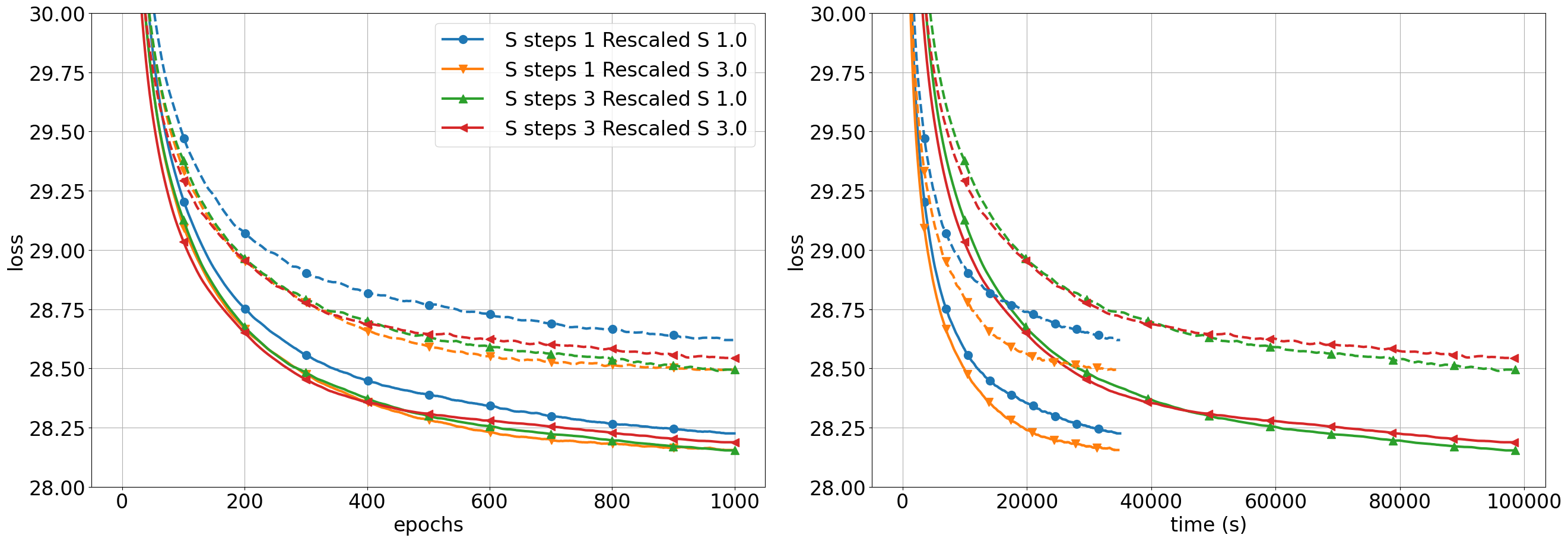}
\caption{Loss on the train and validation sets of the miniMNIST with different 1, 3 sleep steps and 1 or 3 times rescaled sleep steps for NRWS.}\label{fig:SleepWellNRWS}
\end{figure}

Further in Figure \ref{fig:SleepWellNRWS} we studied how the sleep-well affects the NRWS and also its long term affects. We noticed that the NRWS also benefits from extra steps in the sleep cycle, however seemingly to a less degree than the WS. We also tested rescaling the learning rate in the sleep phase, because taking a larger step could benefit the algorithm in a similar way as taking more but smaller steps. We found that slightly increasing the length of the sleep step with a factor of 3 does benefit  NRWS. The benefit of increasing the size rather than the number of steps is that it does not come with any extra time penalty, however taking more steps achieves consistently better minimums. Doing both larger and more steps, however negatively impacts the algorithm (see red line in Figure \ref{fig:SleepWellNRWS}). 

Given the relatively small loss improvement of Figures~\ref{fig:SleepWell} and \ref{fig:SleepWellNRWS}, we decided to opt for the original solution with 1 sleep step and with the same size as the wake phase. This solution is the fastest time-wise and it is inline with all other works from the literature that we compare to. Thus we used the standard single sleep step through all of our experiments, otherwise the improvement from the rescaled sleep step might overshadow the results from the NRWS.

\section{Data augmentation}\label{ss:appDataAugment}

\begin{figure}
\centering
\begin{subfigure}[c]{.33\textwidth}
  \centering
  \includegraphics[width=1.\textwidth]{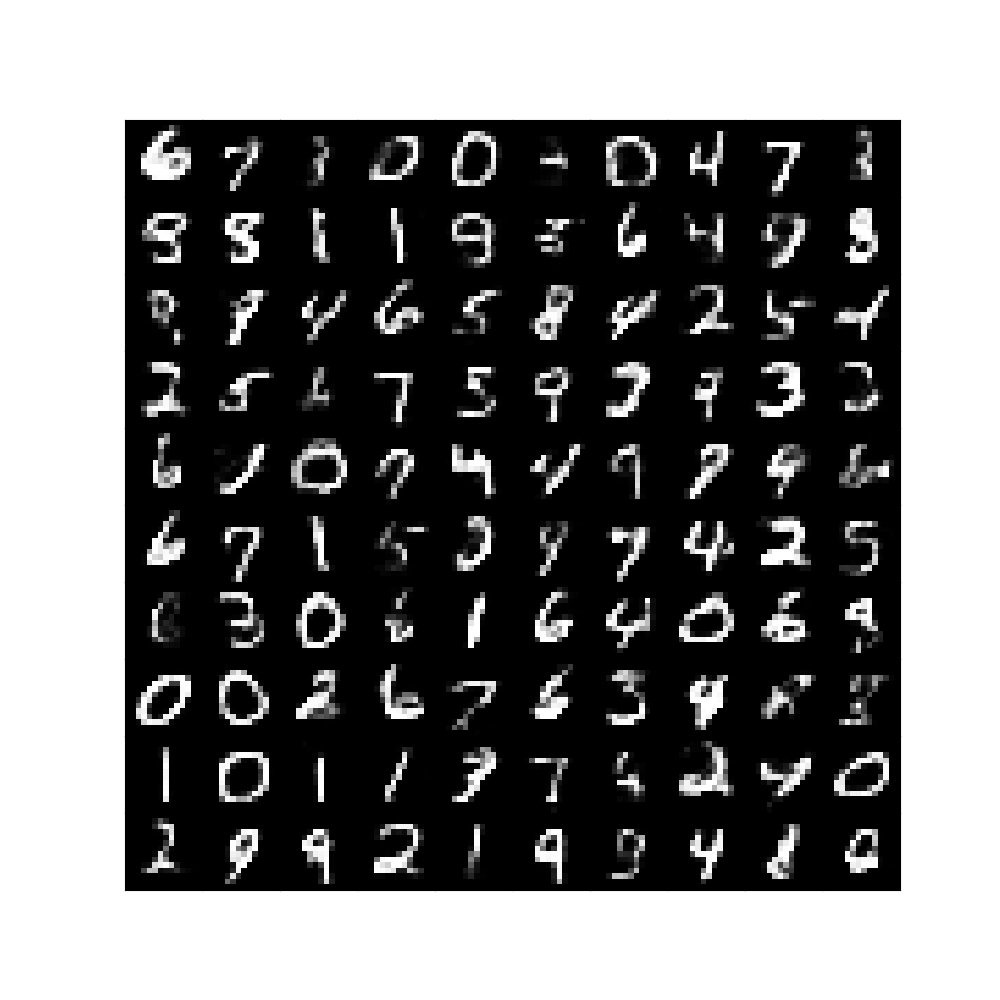}
  \caption{miniMNIST B. probs}
\end{subfigure}%
\begin{subfigure}[c]{.33\textwidth}
  \centering
  \includegraphics[width=1.\textwidth]{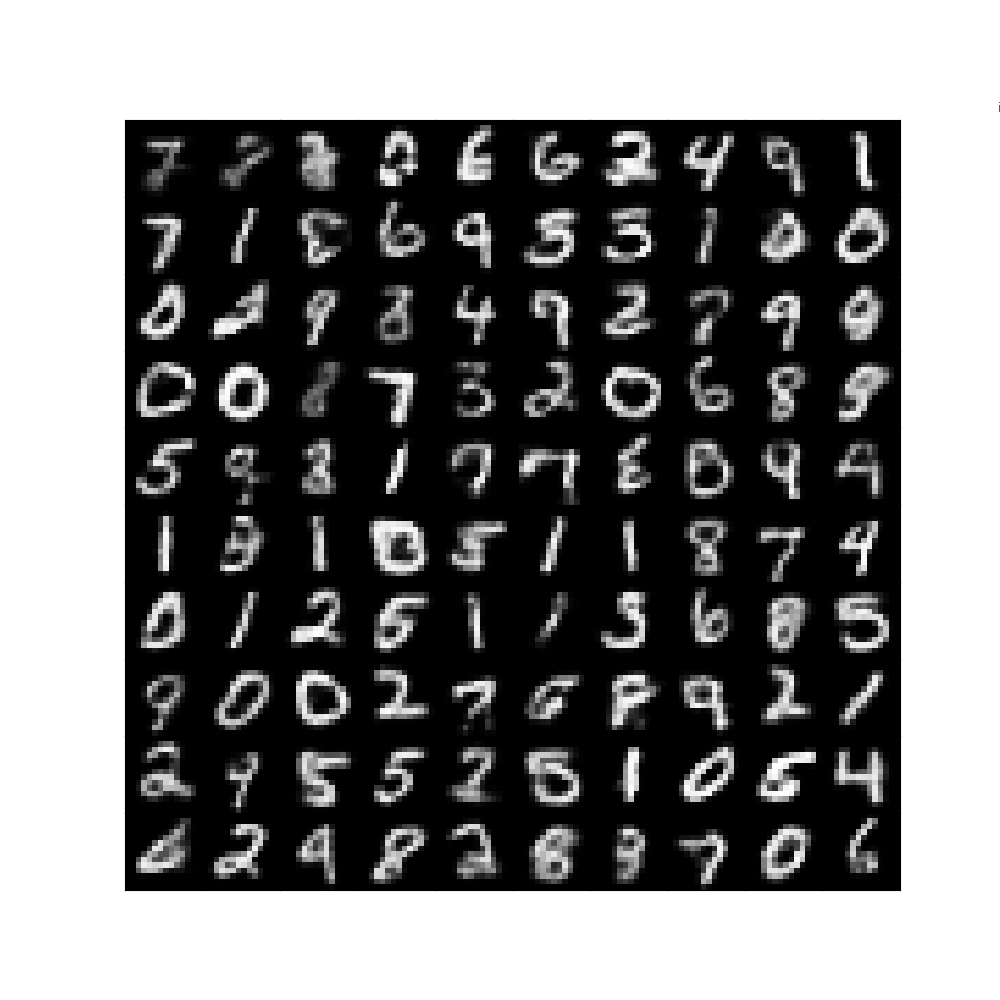}
  \caption{miniMNIST C. probs}
\end{subfigure}%
\begin{subfigure}[c]{.33\textwidth}
  \centering
  \includegraphics[width=1.\textwidth]{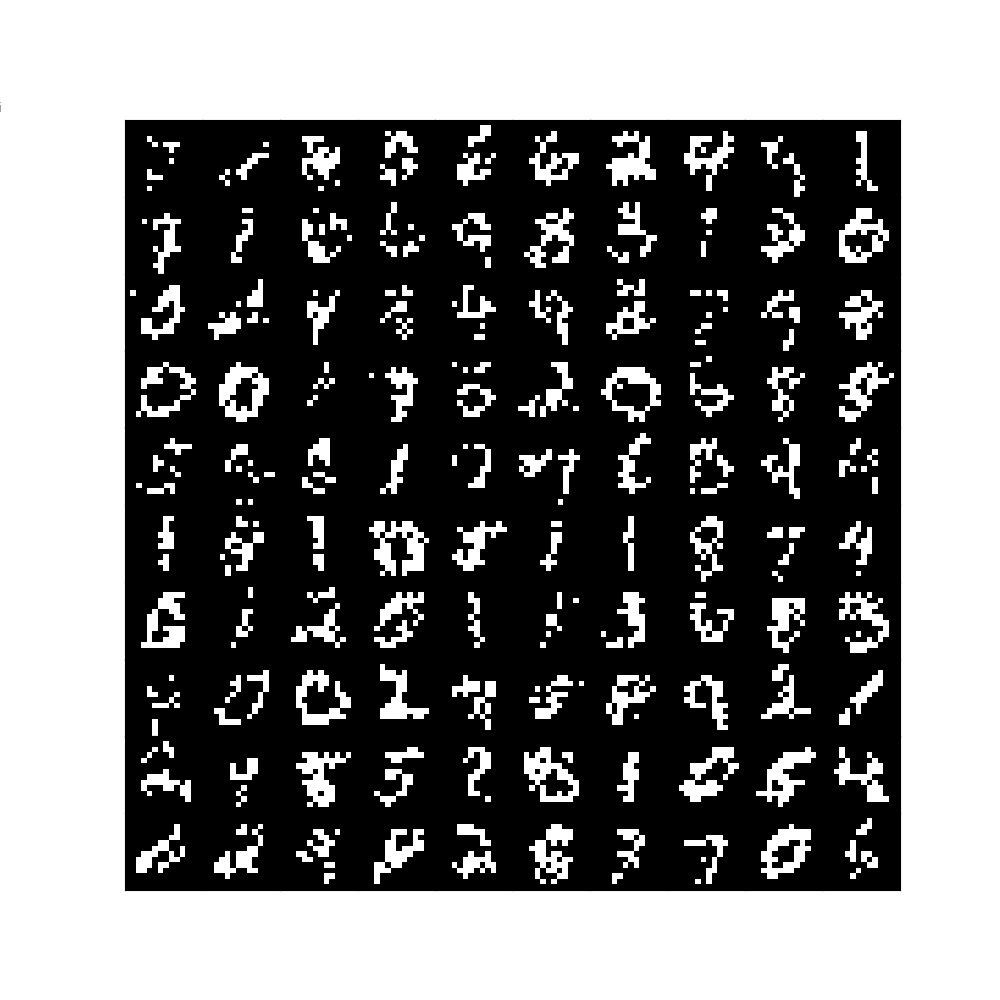}
  \caption{miniMNIST samples}
\end{subfigure}%
\caption{miniMNIST Dataset examples.}
\label{fig:miniMNISTGen}
\end{figure}
 
We compare two different strategies to learn a binary dataset. The first approach is to simply binarize all the samples from the dataset once, by rounding to $\{0,1\}$, which we call \textbf{B}. This technique is used for benchmarking usually and on average have smaller log likelihoods. The second technique, Binary Stochastic or Continuous \textbf{C}, is to take the gray values as the means to a Bernoulli distribution, for each sample from the dataset and for each pixel in the image. This form of data augmentation is enabling us to use continuous data, it is also used commonly in the literature \cite{bornschein_bidirectional_2015}. At each training step we sample from the distributions, thus we get a range of samples, from a single image, which together approximate the original continuous example better than \textbf{B}. In Fig~\ref{fig:miniMNISTGen} \textit{(a)} and \textit{(b)} we see samples from models thought with the differing techniques where we see that the \textbf{C} creates clearly more realistic  images. We use the same technique for TFD in the main paper in Figure \ref{fig:FashionMNISTGen}.

\begin{figure}
\centering
\includegraphics[width=1.0\textwidth]{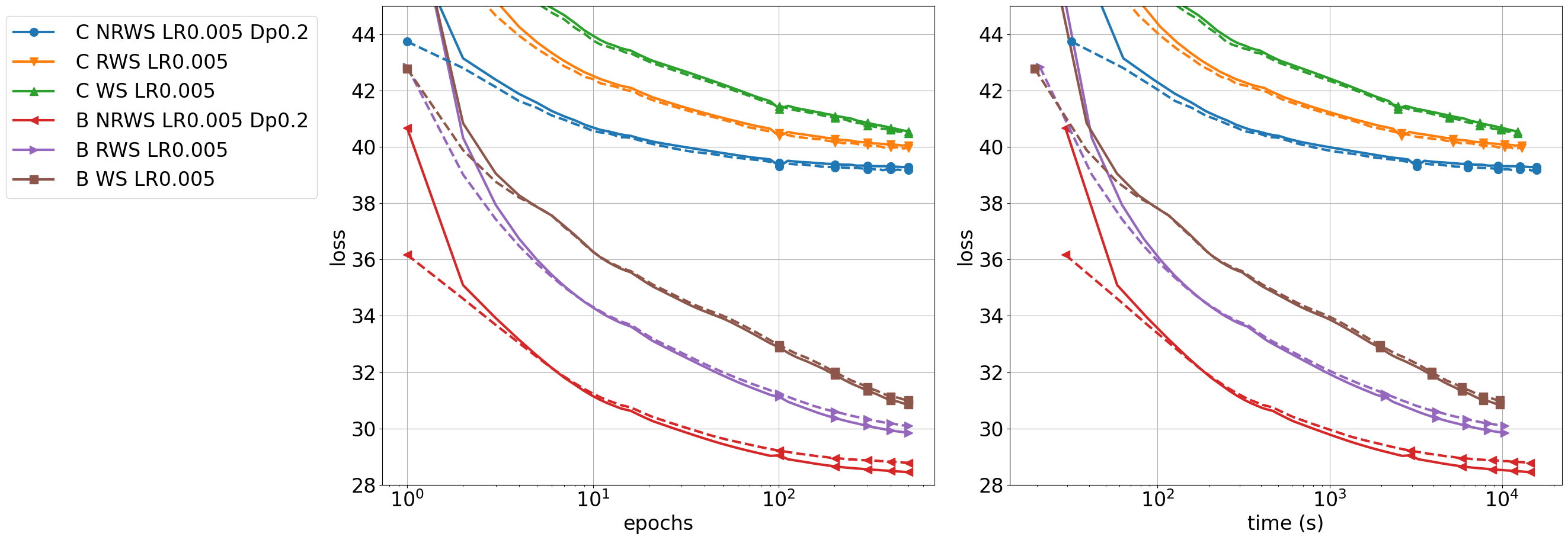}
\caption{The loss of training miniMNIST until convergence with \textbf{B} and \textbf{C}, with the algorithms WS, RWS and NRWS with the layers of the size 100, 50, 20, 10, 10. On the left the convergence in epochs and on the right convergence in wall-clock time (s), both in log-scale [LR=learning rate $\eta$, Dp=Damping factor $\alpha$].}\label{fig:miniMNISTcurves}
\end{figure}

In Figure~\ref{fig:miniMNISTcurves} we compare the loss curves for three different models WS, RWS and NRWS and the two strategies and in Table~\ref{tab:soaminiMNIST} for the miniMNIST dataset, we report the importance sampled approximation of the log-likelihood. We can see that in all cases the NRWS outperforms both models both in achieving the better minimum, as well as in the rate of convergence and wall-clock time, with a clear advantage visible from the very beginning.

\begin{table}
\centering
\begin{tabular}{|c|c|c|c|c|c|c|}
\hline
 \textbf{DS} & \textbf{ALG} & \boldmath{$\eta$} & \textbf{LL}  & \textbf{T/E}  \\
   \hline
 & WS   & 0.04 & -29.337 & 19s  \\
\textbf{B} & RWS  & 0.02 & -28.695 & 21s  \\
 & NRWS & 0.004 & \textbf{-27.606} & 31s  \\
\hline
 & WS  & 0.02 & -38.232 & 23s  \\
\textbf{C} & RWS  & 0.01 & -37.811 & 25s  \\
 & NRWS & 0.004 & \textbf{-36.578} & 32s  \\
\hline
\end{tabular}
\caption{
Importance Sampling estimation of the log-likelihood (\textbf{LL}) with 10,000 samples for different algorithms after 500 epochs of training with SGD. \boldmath{$\eta$} is the learning rate, \textbf{T/E} is the average time per epoch. The damping factor used for NRWS is 0.05. For all algorithms, the number of samples used in training is 10.}\label{tab:soaminiMNIST}
\end{table}

\section{Results for NRWS with Nadam}\label{s:appRes}

As we can see in Table \ref{tab:soaNadam} also when we use Nesterov-Adam (Nadam) instead of SGD (as in the main paper), the NRWS converges to a better minimum, but worse than simple SGD (Table \ref{tab:soa}) with the reasoning mentioned in Section~\ref{ss:soa}.
\begin{table}[ht]
\centering
\begin{tabular}{|c|c|c|c|c|c|}
\hline
 \textbf{ALG} & \textbf{S}  & \boldmath{$\eta$} & \textbf{LL}  & \textbf{T/E}  \\
   \hline
RWS  & 10 & 0.0002 & -86.987 & 34s  \\
NRWS & 10 & 0.001 & \textbf{-85.675} & 44s  \\
   \hline
\end{tabular}
\captionof{table}{
Importance Sampling estimation of the log-likelihood (\textbf{LL}) with 10,000 samples for different algorithms after training till convergence with Nadam. The damping factor used is 0.1. \textbf{T/E} - average time per epoch; \textbf{S} - samples in training, \boldmath{$\eta$} is the learning rate. }\label{tab:soaNadam}
\end{table}

\section{Diagonal Natural Reweighted Wake-Sleep}\label{s:appDNRWS}
The Diagonal Natural Reweighted Wake-Sleep (DNRWS) is a version of the NRWS where we approximate the FIM by taking only its diagonal elements. In the case of the HM it is easy to calculate, instead of performing the calculations in \eqref{eq:estimation-Fisherp} and \eqref{eq:estimation-Fisherq} we can just calculate
\begin{align}
F^{(i)}_{p,j}  &= \frac{1}{n} \sum \sigma^{'}\left(W^{(i)\trasp}_j h^{(i+1)}\right) (h^{(i+1)})^2 \; \text{ and} \\ 
F^{(i)}_{q,j}  &= \frac{1}{n} \sum \sigma^{'}\left(V^{(i)\trasp}_j h^{(i-1)}\right) \, (h^{(i-1)})^2  \; .
\end{align}

Inverting the matrices becomes trivial since the FIM is diagonal, calculating the reciprocal is computationally negligible. In fact because it is much faster, we can calculate the FIM approximation in every gradient step, with no need to save it for $K$ steps.

\begin{figure}[!ht]
\centering
\includegraphics[width=0.5\textwidth]{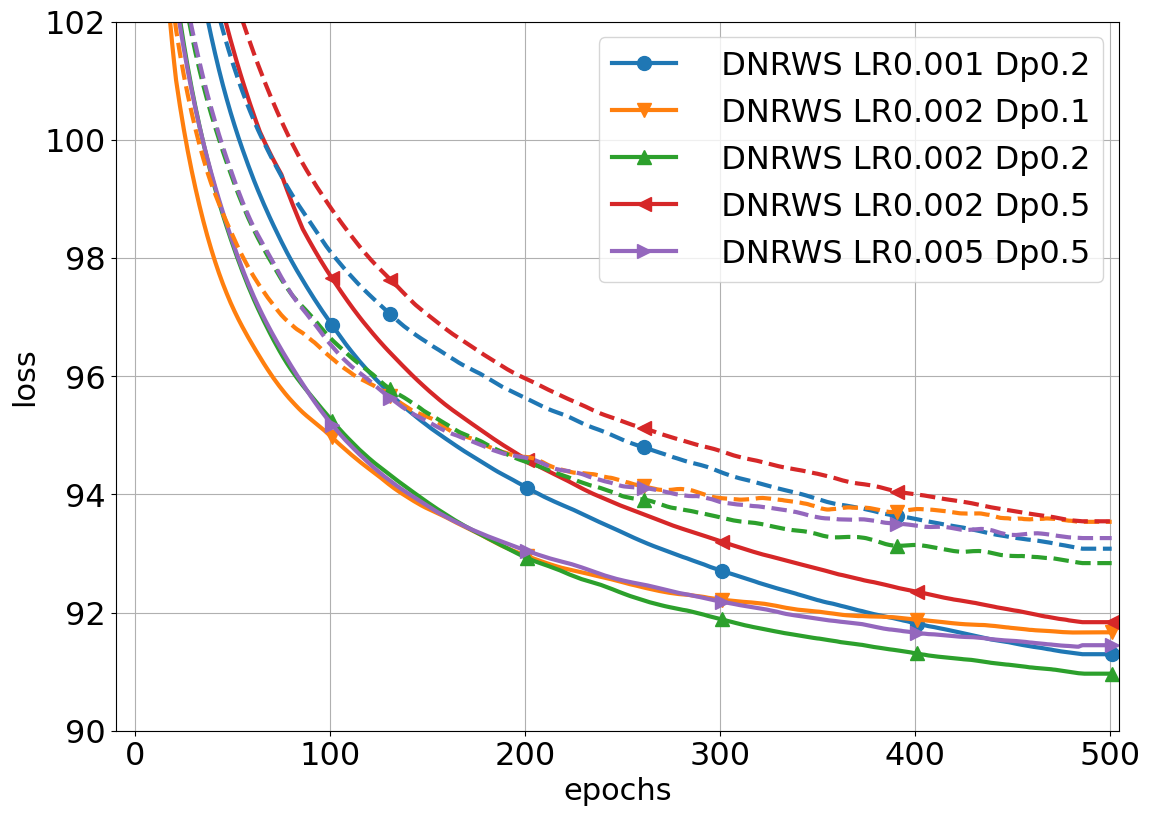}
\caption{Loss of training (continuous line) and validation (dashed line) on MNIST with different Learning Rates and Damping Factors for the DNRWS [LR=learning rate $\eta$, Dp=Damping factor $\alpha$].}\label{fig:DNRWS}
\end{figure}

\begin{figure}[!ht]
\centering
\includegraphics[width=1.0\textwidth]{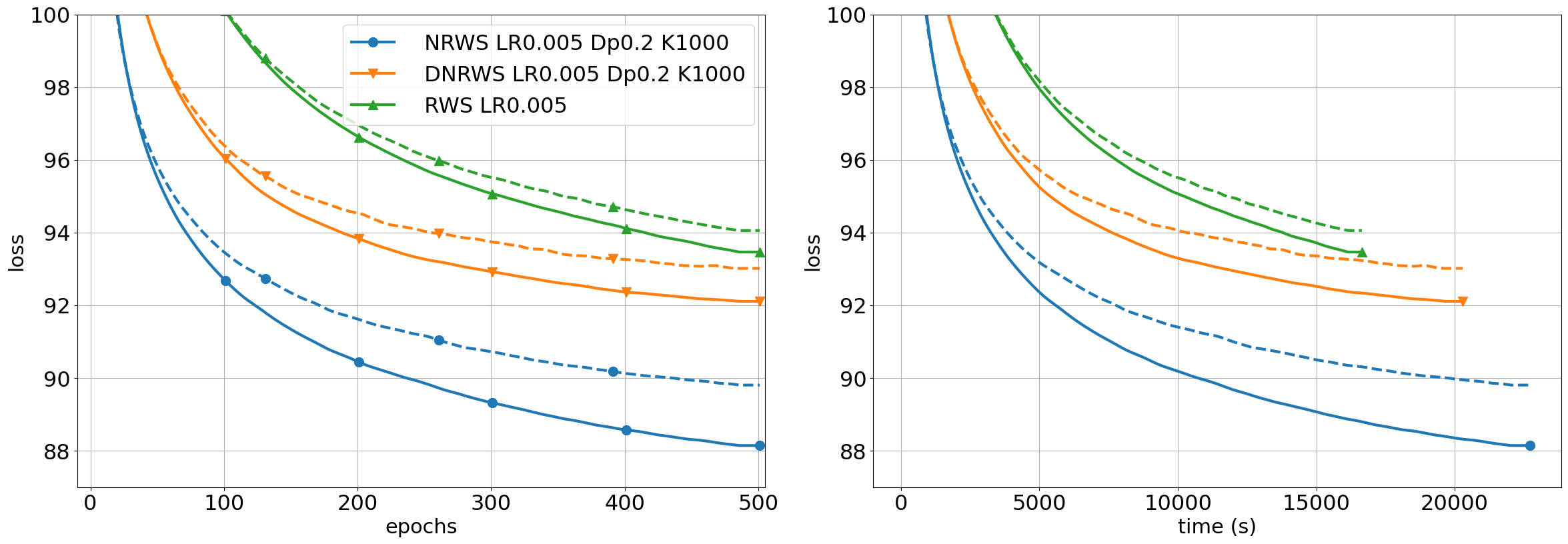}
\caption{Loss of training (continuous line) and validation (dashed line) of RWS, DNRWS and NRWS on MNIST; (right) epochs (left) seconds of 500 epochs [LR=learning rate $\eta$, Dp=Damping factor $\alpha$, K=K-step].}\label{fig:DNRWS_comp}
\end{figure}

Analyzing how the change in the hyperparameters of the learning rate and damping factor $\alpha$ in Figure \ref{fig:DNRWS} it reveals that usually the best combination is similar to the one used for NRWS. Taking a smaller $\alpha$ leads to quicker convergence, but worse minimum, with some instability when closer to convergence. Larger damping leads to a more stable convergence, but a slower one, compensating by speeding up with a larger learning rate leads to premature convergence.

In Figure \ref{fig:DNRWS_comp} we compare the DNRWS to NRWS and RWS for a shorter period of 500 epochs where every hyperparameter like sample and minibatch-size was kept the same for all algorithms, with a full comparison till convergence can be found in Section~\ref{sec:experiments} in the main paper. We see a speedup of the DNRWS compared to the RWS, but the achieved minimum is worse than that of the NRWS. This observation is in line with what we were expecting as the diagonal approximation of the FIM leads to worse results than the estimation of the actual structure. 

\end{appendices}

\end{document}